\documentclass{article} % For LaTeX2e
\usepackage{paper,times}

\usepackage{amsmath,amsfonts,bm}

\def\eqref#1{equation~\ref{#1}}
\def\1{\bm{1}}

\DeclareMathAlphabet{\mathsfit}{\encodingdefault}{\sfdefault}{m}{sl}
\SetMathAlphabet{\mathsfit}{bold}{\encodingdefault}{\sfdefault}{bx}{n}

\newcommand{\R}{\mathbb{R}}

\DeclareMathOperator*{\argmin}{arg\,min}

\usepackage{hyperref}
\usepackage{enumitem}
\usepackage{tocloft}
\usepackage{xcolor}
\usepackage{graphicx} 
\usepackage{array}
\usepackage{amssymb}
\usepackage{booktabs}
\usepackage{wrapfig}
\usepackage{ulem}
\usepackage{tikz}
\usepackage{subcaption}
\usepackage{colortbl}
\usepackage{tcolorbox}
\usepackage{algorithm2e}
\usepackage{multirow}
\definecolor{customorange}{HTML}{FF7F0E}
\definecolor{customred}{HTML}{D62728}
\definecolor{customgreen}{HTML}{009955}
\definecolor{customgray}{HTML}{E8E8E8}
\definecolor{gray}{rgb}{0.95,0.95,0.95}
\usepackage{listings}
\colorlet{punct}{red!60!black}
\definecolor{background}{HTML}{EEEEEE}
\definecolor{delim}{RGB}{20,105,176}
\colorlet{numb}{magenta!60!black}
\usepackage[mode=match]{siunitx}
\usepackage{inconsolata}  % Nicer \texttt font
 \usepackage{hyperref}
\usepackage[capitalize]{cleveref}  % Automatic labelling of cross-references, e.g. "\cref{tab:foo}" -> "Table~1"
\lstdefinelanguage{json}{
    basicstyle=\normalfont\ttfamily\footnotesize,
    numbers=left,
    numberstyle=\scriptsize,
    stepnumber=1,
    numbersep=8pt,
    showstringspaces=false,
    breaklines=true,
    frame=single,
    backgroundcolor=\color{background},
    literate=
     *{0}{{{\color{numb}0}}}{1}
      {1}{{{\color{numb}1}}}{1}
      {2}{{{\color{numb}2}}}{1}
      {3}{{{\color{numb}3}}}{1}
      {4}{{{\color{numb}4}}}{1}
      {5}{{{\color{numb}5}}}{1}
      {6}{{{\color{numb}6}}}{1}
      {7}{{{\color{numb}7}}}{1}
      {8}{{{\color{numb}8}}}{1}
      {9}{{{\color{numb}9}}}{1}
      {:}{{{\color{punct}{:}}}}{1}
      {,}{{{\color{punct}{,}}}}{1}
      {\{}{{{\color{delim}{\{}}}}{1}
      {\}}{{{\color{delim}{\}}}}}{1}
      {[}{{{\color{delim}{[}}}}{1}
      {]}{{{\color{delim}{]}}}}{1},
}

\title{An X-Ray Is Worth 15 Features: Sparse Autoencoders for Interpretable Radiology Report Generation}

\author{
  Ahmed Abdulaal$^{1}$\thanks{Joint first authors.} \quad Hugo Fry$^{2*}$ \quad Nina Monta\~na-Brown$^{1}$ \quad Ayodeji Ijishakin$^{1}$ \\ 
  \ \textbf{Jack Gao}$^{1}$ \quad \textbf{Stephanie Hyland}$^{3}$ \quad \textbf{Daniel C. Alexander}$^{1}$ \quad \textbf{Daniel C. Castro}$^{3}$ \\ [2mm]
  $^1$Hawkes Centre, UCL, London, UK \\
  $^2$Independent Researcher \\
  $^3$Microsoft Research, Cambridge, UK \\
}

\iclrfinalcopy % Uncomment for camera-ready version, but NOT for submission.
\begin{document}

\maketitle

\begin{abstract}
Radiological services are experiencing unprecedented demand, leading to increased interest in automating radiology report generation. Existing Vision-Language Models (VLMs) suffer from hallucinations, lack interpretability, and require expensive fine-tuning. We introduce SAE-Rad, which uses sparse autoencoders (SAEs) to decompose latent representations from a pre-trained vision transformer into human-interpretable features.
Our hybrid architecture combines state-of-the-art SAE advancements, achieving accurate latent reconstructions while maintaining sparsity.
Using an off-the-shelf language model, we distil ground-truth reports into radiological descriptions for each SAE feature, which we then compile into a full report for each image, eliminating the need for fine-tuning large models for this task. To the best of our knowledge, SAE-Rad represents the first instance of using mechanistic interpretability techniques explicitly for a downstream multi-modal reasoning task. On the MIMIC-CXR dataset, SAE-Rad achieves competitive radiology-specific metrics compared to state-of-the-art models while using significantly fewer computational resources for training. Qualitative analysis reveals that SAE-Rad learns meaningful visual concepts and generates reports aligning closely with expert interpretations. Our results suggest that SAEs can enhance multimodal reasoning in healthcare, providing a more interpretable alternative to existing VLMs.
\end{abstract}

\section{Introduction}
Radiological services are essential to modern clinical practice, with demand rising rapidly. In the UK, the NHS performs over 43 million radiological procedures annually \citep{lewis2021pathology}, costing over £2 billion, and demand for scans more than doubled between 2012 and 2019 \citep{england2019transforming}. A significant portion of these costs addresses rising demand through agency, bank, and overtime staff, but a national imaging strategy notes this funding is unsustainable \citep{england2019transforming}. Consequently, there's growing interest in (semi)-automating tasks like radiology report generation, augmentation, and summarization to assist clinicians \citep{zhu2024leveraging, chen2024chexagent, perez2024rad}, spurred by advances in multimodal text-vision modelling techniques.

Recent architectures that combine vision encoders with pretrained Large Language Models (LLMs) to create multimodal Vision-Language Models (VLMs) have shown impressive performance in visual and language tasks \citep{liu2024visual, liu2023improved, li2024llava, lin2023video, liu2023llava}. VLMs have been applied to healthcare tasks, including radiology report generation \citep{hyland2023maira, bannur2024maira, chen2024chexagent, stock2024generalist, yang2024advancing}, typically by mapping image representations into the LLM's token embedding space. The LLM is fine-tuned to respond to prompts like `\textit{$<$image tokens$>$ Produce the findings section of a radiology report for this image}' \citep{perez2024rad}. Despite improvements from scaling VLMs \citep{yang2024advancing}, hallucinations and disagreements with domain experts remain common \citep{yildirim2024multimodal, jeblick2023chatgpt, lee2023benefits, tanno2024consensus}.
\begin{figure}[!ht] 
    \centering
    \includegraphics[width=0.9\linewidth]{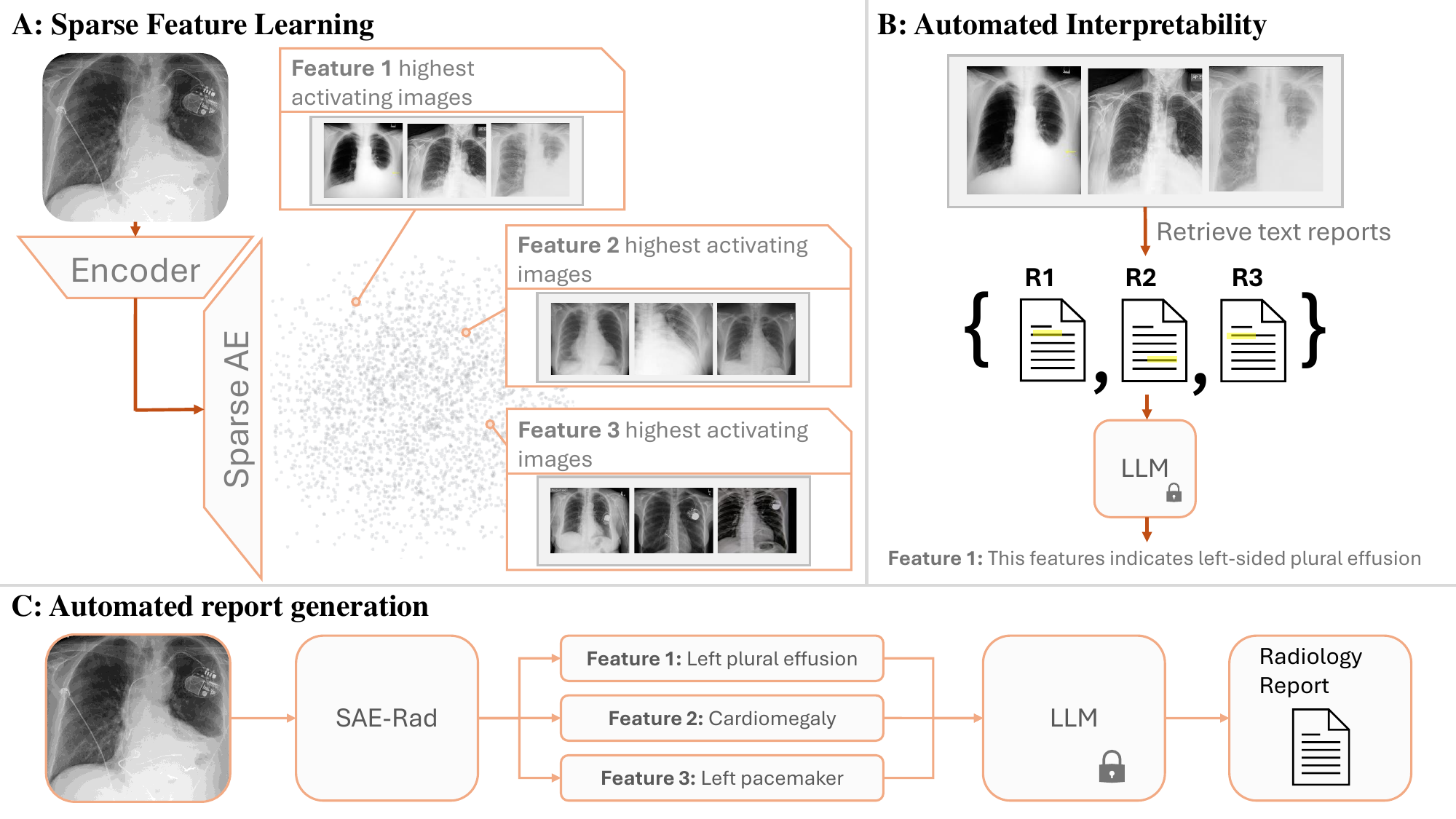}
    \caption{\textbf{SAE-Rad overview}. \textbf{Panel A:} We learn a set of sparsely activating features by training a Sparse Autoencoder (SAE) on class tokens produced by a radiology-image encoder. \textbf{Panel B:} We retrieve the corresponding reference reports for highest activating images for a feature, from which we can produce text descriptions of each feature. \textbf{Panel C:} We pass a new image through the radiology-image encoder and SAE encoder to retrieve the highest activating features. Text descriptions of these features are subsequently used by a pretrained large language model (LLM) to generate a detailed radiology report.}
    \label{fig:experimental_setup}
\end{figure}
Hallucinations are unavoidable in LLMs \citep{kalai2024calibrated, xu2024hallucination}, and whilst this represents a limitation of current VLM systems designed for radiology report generation, there are other important considerations of using such a system for this task. For current state-of-the-art systems, it is necessary to finetune a multi-billion parameter LLM (as well as projector weights) to perform visual instruction tuning \citep{liu2024visual}, which is computationally intensive and can be prohibitively expensive. Additionally, the generated reports a VLM provides may not be faithful to the underlying computations of the image encoder -- we should aim to design a framework which is verifiably faithful to the image model by reverse engineering the computations of the image encoder. This could yield more interpretable results and thus engender more trust in automated radiology reporting systems. 

To this end, we introduce SAE-Rad (\cref{fig:experimental_setup}), a framework which leverages mechanistic interpretability techniques, and in particular sparse autoencoders (SAEs), to directly decompose image class tokens from a pre-trained radiology image encoder \citep{perez2024rad} into human-interpretable features. These interpretable features are labelled by use of pre-trained LLMs in an automated interpretability pipeline \citep{bricken2023towards}, and can then be compiled into descriptive radiology reports. To the best of our knowledge, this is the first instance of using SAEs explicitly for a downstream task requiring multi-modal reasoning. Our contributions are as follows:
1)~\textit{Novel SAE:} We propose a hybrid architecture which builds on gated SAEs to achieve accurate reconstructions with comparable sparsity to state-of-the-art techniques;
2)~\textit{Multimodal reasoning pipeline:} We develop a framework based on our SAE which leverages automated interpretability to label SAE features using paired text data and ultimately produce radiology reports given an image;
3)~\textit{Mechanistic interpretability in healthcare:} Our case study on healthcare data demonstrates the utility of mechanistic interpretability for downstream tasks in a relatively smaller and significantly more homogeneous dataset than most general-domain text or natural-image datasets.    

\section{Related work}
\paragraph{Multimodal reasoning} Multimodal reasoning methods like ScienceQA \citep{lu2022learn} introduced multimodal chain-of-thought (CoT) by zero-shot prompting models to generate rationales and answers simultaneously. Multimodal-CoT (MM-CoT) \citep{zhang2023multimodal} extended this with a two-stage framework that separates rationale generation and answer inference using two models of the same architecture. Duty-Distinct CoT (DDCoT) \citep{zheng2023ddcot} further factorizes rationales by decomposing the initial question into sub-questions answered by a vision-language model (VLM). Other divide-and-conquer approaches decompose questions into sub-questions but often require training task-specific visual question generation (VQG) models and additional scoring models \citep{selvaraju2020squinting, uehara2022learning, wang2022co}. IdealGPT \citep{you2023idealgpt} iteratively decomposes queries and uses a VLM to answer sub-questions, repeating the process if confidence is low. Unlike these methods that rely on decomposing questions or generating rationales through additional models, our approach directly extracts and interprets features from pre-trained image encoders. This enables faithful and transparent multimodal reasoning without the need for extensive fine-tuning or supplementary VQG models.

\paragraph{Radiological VLMs}  A number of works have finetuned or otherwise trained specialised foundation models for radiological applications including Med-flamingo~\citep{moor2023med}, Med-PaLM M~\citep{tu2024towards}, LLava-Med~\citep{li2024llava}, Med-Gemini~\citep{yang2024advancing}, Rad-DINO~\citep{perez2024rad}, MAIRA-1~\citep{hyland2023maira}, R2gengpt~\citep{wang2023r2gengpt}, and Radiology-GPT~\citep{liu2023radiology}. With regards to radiology report generation, several works have focused on producing both the `findings' and `impression' sections of the reports \citep{chen2020generating, jin2024promptmrg, yan2023style}, whilst others have focussed on the `impression' section \citep{bannur2023biovilt}, or, most commonly, the `findings' section \citep{tu2024towards, miura2020improving, delbrouck2022improving, tanida2023interactive, nicolson2023improving}. As noted by others \citep{hyland2023maira, yu2023evaluating, jeong2024multimodal}, studies examining all three settings found that the choice of section(s) to report significantly affects the performance metrics, making comparison between results difficult. For this reason we focus on the most common setting of producing the `findings' section of a radiology report. 

\paragraph{Mechanistic interpretability} \citet{bricken2023towards} demonstrated that SAEs could recover monosemantic features by training on the residual stream of small transformers. The gated SAE was a Pareto improvement over the baseline SAE in terms of sparsity as measured by the L0 and the loss recovered \citep{rajamanoharan2024improving}. There was concern that SAEs would not scale to frontier transformers until recent work by \citet{templeton_scaling_2024} which trained SAEs on Claude 3 Sonnet and discovered a large number of monosemantic features. Contemporaneously, SAEs were trained on the class tokens of a CLIP vision transformer \citep{fry2024multimodal}, InceptionV1 \citep{gorton2024missing}, the conditioning embeddings of diffusion models \citep{goh2024clip}, and the vision transformer of a pathology foundation model \citep{le2024interpretability}. However, in all cases the discovered language/visual features were not used to perform a downstream multimodal task. 

\section{Background}
In this section we give a brief overview of mechanistic interpretability, SAEs, and gated SAEs, before introducing our SAE-Rad framework. 
\subsection{Mechanistic interpretability and sparse autoencoders (SAEs)}
\paragraph{Mechanistic interpretability} Mechanistic interpretability research aims to identify, understand, and verify the algorithms that an ML model implements by reverse engineering a model's computations into human-interpretable components \citep{olah2020zoom, rajamanoharan2024improving}. Classical approaches attempted to achieve this by analysing the firing patterns of individual neurons, which were interpreted as possible `concept representations’. However, this was broadly ineffective as neurons can be \textit{polysemantic}, meaning that a single neuron may fire on many unrelated concepts \citep{rajamanoharan2024improving, bolukbasi2021interpretability, Elhage2022}. Polysemantic neurons are believed to arise during training due to the composition of both the linear representation and superposition hypotheses. 
\paragraph{Linear representation and superposition hypotheses} Motivated by a number of findings which suggest that concept representations are linear \citep{gurnee2023finding, olah2020zoom, park2023linear}, the linear representation hypothesis states that neural networks represent concepts (sometimes interchangeably referred to as `features') as directions in \textit{activation space} \citep{Nanda2024}. This hypothesis can be thought of as being composed of two properties: 1) Linearity: That is, features are represented as directions; and 2) Decomposability: We can understand neural network outputs as a composition of multiple independently understandable features \citep{Elhage2022Toy}. The superposition hypothesis states that for an intermediate representation of dimension $n$, neural networks will encode $M\gg n$ concepts as linear directions \citep{rajamanoharan2024improving, Elhage2022Toy}. These directions form an overcomplete basis of the activation space and must therefore necessarily overlap with each other. However, a single input will only activate a sparse subset of these concepts, leading to minimal interference between the (non-orthogonal) concept directions \citep{gurnee2023finding, rajamanoharan2024improving}. The superposition hypothesis can be thought of as a form of learned neural network compression and is closely related to compressed sensing. Recent work \citep{bricken2023towards} has proposed using SAEs to take features out of superposition and learn monosemantic interpretable representations.
\paragraph{Sparse autoencoders (SAEs)} SAEs attempt to ‘undo’ superposition by learning the sparse overcomplete basis \citep{mallat1993matching, rajamanoharan2024improving}  (or dictionary) of the activation space induced by superposition. SAEs attempt to learn both the concept directions and a sparse vector of coefficients for the inputs, that reflect how much each concept is activated for each input \citep{cunningham2023sparse, bricken2023towards}. To align our nomenclature with the recent literature, we will henceforth refer to such sparse vectors of coefficients as `feature activations'.  

We begin by defining the `baseline SAE' described by \citet{bricken2023towards}. Let $n$ be the dimension of the input and output (typically the input is the residual stream of a transformer, and the output is its reconstruction \citep{transformer_framework}), and $m$ be the SAE hidden layer dimension. Let $s$ be the size of the dataset. Then given encoder weights and biases $W^\text{enc}\in\R^{m\times n}, \mathbf{b}^\text{enc}\in\R^m$, and decoder weights and biases $W^\text{dec}\in\R^{n\times m}, \mathbf{b}^\text{dec}\in\R^n$, the encoding and decoding operations for a dataset $X\in\R^{s,n}$ are
\begin{align}
    \mathbf{h}(\mathbf{x})&:=\text{ReLU}(W^\text{enc}(\mathbf{x} - \mathbf{b}^{\text{dec}}) + \mathbf{b}^\text{enc})\label{eq:vanilla_sae_enc}\\
    \hat{\mathbf{x}}(\mathbf{h}(\mathbf{x}))&:=W^\text{dec}\mathbf{h}(\mathbf{x}) + \mathbf{b}^\text{dec}. \label{eq:decoder}
\end{align}
The loss function is then
\begin{equation}\label{eqn:eq_loss_naive_sae}
    \mathcal{L}(\mathbf{x}):=\frac{1}{|X|}\sum_{\mathbf{x}\in X}\left[\|\mathbf{x}-\mathbf{\hat{x}}(\mathbf{h}(\mathbf{x}))\|_2^2+\lambda\|\mathbf{h}(\mathbf{x})\|_1\right],
\end{equation}
where $\lambda$ is an L1 sparsity coefficient. The first term is a reconstruction error measured by the squared distance between the input and its reconstruction, and the second is an L1 regularization loss to induce sparsity. It should be noted that in this regime it is possible to reduce the second term in \cref{eqn:eq_loss_naive_sae} by simply decreasing the norm of the encoder weights $W^{\text{enc}}$. It is possible to retain both reconstruction quality and sparsity by a corresponding increase of norm of the decoder weights \citep{bricken2023towards, rajamanoharan2024improving}. This effect is not desired due to both overflow (decoder norm) and underflow (encoder norm) errors. Additionally, this can cause unstable training when using adaptive optimization algorithms such as Adam \citep{kingma2014adam}\footnote{This is because the Adam algorithm varies the learning rate according to the reciprocal of the exponential moving average of the mean-square gradient.}. To resolve this, the column-wise norm of the decoder $W^{\text{dec}}$ can be constrained during training \citep{bricken2023towards}.

\subsection{Gated SAEs}\label{sec:gated_sae_desc}
As can be seen in \cref{eqn:eq_loss_naive_sae}, SAEs jointly optimize two opposing objectives: 1)~Reconstruction fidelity and 2)~L1 regularization as a proxy for sparsity (as measured by L0). This means the SAE is free to trade-off some reconstruction fidelity in order to perform better on the sparsity penalty. One consequence of this is shrinkage \citep{wright2024addressing}. That is, for a fixed decoder, the sparsity penalty pushes the feature activations $\mathbf{h}(x)$ towards zero whilst the squared distance loss encourages $\mathbf{h}(x)$ to be large enough in order to produce high quality reconstructions. Thus, the standard SAE will systematically underestimate the optimal magnitude of feature activations (and simply rescaling these does not necessarily overcome this bias) \citep{rajamanoharan2024improving, wright2024addressing}. 

\citet{rajamanoharan2024improving} propose a gated SAE which separates the encoding procedure into two tasks: 1) Detecting which features should activate for a given input (this requires an L1 penalty if the features are to be sparse); and 2) Estimating the magnitude of the feature activations (this does \textbf{not} require an L1 loss; indeed, including this loss here introduces a shrinkage bias). The architecture of the gated encoder is 
\begin{equation}
    \mathbf{h}(\mathbf{x}):=\underbrace{\mathbb{I}[\overbrace{W^{\text{gate}}(\mathbf{x}-\mathbf{b}^{\text{dec}})+\mathbf{b}_{\text{gate}}}^{\boldsymbol{\pi}_{\text{gate}}(\mathbf{x})}>0]}_{\mathbf{h}_{\text{gate}}(\mathbf{x})}\odot\underbrace{\text{ReLU}(W^{\text{mag}}(\mathbf{x}-\mathbf{b}^{\text{dec}})+\mathbf{b}_{\text{mag}})}_{\mathbf{h}_{\text{mag}}(\mathbf{x})},
\end{equation}
where $\mathbb{I}[\bullet > 0]$ is an element-wise Heaviside step function and $\odot$ is element-wise multiplication. The $\mathbf{h}_{\text{gate}}$ sub-function learns which features should activate for a given input and $\mathbf{h}_{\text{mag}}$ estimates the magnitude of activations for these features. Here, $\boldsymbol{\pi}_{\text{gate}}$ is referred to as the  $\mathbf{h}_{\text{gate}}$ sub-function's `pre-activations'. To minimize the number of additional parameters required, $W^{\text{mag}}$ shares the same feature directions as $W^{\text{gate}}$, and is defined as
$
    W^{\text{mag}}_{i,j}:=\exp(\mathbf{r}^{\text{mag}}_i)\cdot W^{\text{gate}}_{i,j}
$,
where $\mathbf{r}^{\text{mag}}\in\R^m$ is a vector-valued scaling parameter.
Letting $\text{RA}(\cdot):=\text{ReLU}(\boldsymbol{\pi}_{\text{gate}}(\cdot))$ denote the rectified pre-activations of the gating sub-function, the loss function is defined as
\begin{equation}
    \mathcal{L}(\mathbf{x}):=\underbrace{\|\mathbf{x}-\hat{\mathbf{x}}(\mathbf{h}(\mathbf{x}))\|_2^2}_{\mathcal{L}_{\text{reconstruct}}}+\underbrace{\lambda\|\text{RA}(\mathbf{x})\|_1}_{\mathcal{L}_{\text{sparsity}}}+\underbrace{\|\mathbf{x}-\hat{\mathbf{x}}_{\text{frozen}}(\text{RA}(\mathbf{x}))\|_2^2}_{\mathcal{L}_{\text{aux}}},
\end{equation}
where $\hat{\mathbf{x}}_{\text{frozen}}$ is a fixed copy of the decoder so that gradients from the auxiliary loss $\mathcal{L}_{\text{aux}}$ do not back-propagate to the decoder weights or bias terms. The auxiliary term $\mathcal{L}_{\text{aux}}$ ensures that $\mathbf{h}_{\text{gate}}$ correctly identifies features necessary for reconstruction, as its (positive) pre-activations must be able to reproduce the input. The sparsity term $\mathcal{L}_{\text{sparsity}}$ applies an L1 penalty to the rectified pre-activations (and thus sparsity is only imposed on the gating sub-function), and the reconstruction term serves the same function as in \cref{eqn:eq_loss_naive_sae}.
\section{SAE-Rad}
In this section we introduce SAE-Rad. First, we describe the autoencoder architecture, which is based on the gated SAE described in \cref{sec:gated_sae_desc}. Then, we describe our end-to-end radiology report generation pipeline. 
\subsection{SAE architecture}\label{sec:sae_rad_sae_arch}
It was recently demonstrated that a lower overall SAE loss is achievable without constraining the L2 norm of the decoder weights, or centering the input based on the decoder bias \citep{transformer2024}. Concretely, they use the same decoder as in \cref{eq:decoder} and define the encoder as
\begin{align}
    \mathbf{h}(\mathbf{x})&:=\text{ReLU}(W^\text{enc}\mathbf{x} + \mathbf{b}^\text{enc}).
\end{align} 
Note that the input $\mathbf{x}$ is no longer centered by subtracting the decoder bias $\mathbf{b}^{\text{dec}}$ as in \cref{eq:vanilla_sae_enc}. The sparsity penalty in the loss also now includes the L2 norm of the columns of the decoder $W^{\text{dec}}$:
\begin{equation}\label{eqn:eq3}
    \mathcal{L}(\mathbf{x}):=\|\mathbf{x}-\mathbf{\hat{x}}(\mathbf{h}(\mathbf{x}))\|_2^2+\lambda\sum_i\mathbf{h}_i(\mathbf{x})\cdot\|W_{\cdot,i}^\text{dec}\|_2.
\end{equation}
The feature activation for a feature $i$ is then $\mathbf{h}_i(\mathbf{x})\cdot\|W_{\cdot,i}^{\text{dec}}\|_2$. The `concept directions' are the unit-normalized decoder vectors $\frac{W_{\cdot,i}^{\text{dec}}}{\|W_{\cdot,i}^{\text{dec}}\|_2}$.

The SAE-Rad sparse autoencoder is a hybrid architecture which combines a gated encoder layer with unconstrained decoder norms. Its encoder is defined as 
\begin{align}
    \mathbf{h}(\mathbf{x})&:=\mathbb{I}[W^{\text{gate}}\mathbf{x}+\mathbf{b}^{\text{gate}}>0]\odot\text{ReLU}(W^{\text{mag}}\mathbf{x}+\mathbf{b}^{\text{mag}}),
\end{align}
and the training objective given by
\begin{equation}\label{eqn:SAE_RAD_eqns}
    \mathcal{L}(\mathbf{x}):=\underbrace{\|\mathbf{x}-\hat{\mathbf{x}}(\mathbf{h}(\mathbf{x}))\|_2^2}_{\mathcal{L}_{\text{reconstruct}}}+\underbrace{\lambda\sum_i\text{RA}_i(\mathbf{x})\cdot\|W^{\text{dec}}_{\cdot, i}\|_2}_{\mathcal{L}_{\text{sparsity}}}+\underbrace{\|\mathbf{x}-\hat{\mathbf{x}}(\text{RA}(\mathbf{x}))\|_2^2}_{\mathcal{L}_{\text{aux}}}.
\end{equation}
There are therefore four architectural differences between the gated SAE and SAE-Rad: 1) The L2 norm of the decoder weights are not constrained to unit norm; 2) The L2 norm of the decoder weights are included in the sparsity loss term; 3) The input is not centered by subtracting the decoder bias term; 4) We do not leverage stop gradient operations/decoder copying -- this means that we allow gradients to back-propagate to the decoder parameters from the auxiliary loss. SAEs are trained to maximize reconstruction fidelity for a given sparsity level \citep{rajamanoharan2024improving}. We therefore evaluate SAEs based on the L0 norm, which is defined as the average number of firing features for the inputs $\mathbb{E}_{\mathbf{x}\sim X}\|\mathbf{h}(x)\|_0$, and the mean-squared error loss as defined in \cref{eqn:eq_loss_naive_sae}. We demonstrate that this novel SAE architecture outperforms the architecture described by \citet{transformer2024} on our dataset, achieving both a lower L0 and a lower mean-squared error. Details of this comparison can be found in \cref{subsec:architecture_comparison}.

\subsection{SAE-Rad -- Automated radiology reporting pipeline}\label{sec:sae_rad_full_pipeline}
In this section we describe our pipeline to automate the task of radiology report generation. Concretely, a radiographic image $\mathbf{x}$ is passed through a pre-trained and frozen vision encoder $f_{\text{img}}(\cdot):\mathbf{x}\mapsto\mathbf{z}$ to produce an image latent $\mathbf{z}$. We leverage the hybrid SAE architecture described in \cref{sec:sae_rad_sae_arch} to learn feature directions $\frac{W_{\cdot,i}^{\text{dec}}}{\|W_{\cdot,i}^{\text{dec}}\|_2}$ and their associated activations $\mathbf{h}(\mathbf{z})\cdot\|W^{\text{dec}}\|_2$ from the latents.

In order to generate a text-based report, we produced plain-English descriptions of the learnt sparse dictionary (i.e., a description of what each feature direction represents). To do this we performed automated interpretability \citep{bricken2023towards} by using a pre-trained and frozen LLM to analyse the ground-truth radiology reports of the highest activating images for each feature. Let $\mathbf{X}_{\text{highest}}^{(i)}$ represent the set of images with the largest feature activations for feature $i$. For each image $\mathbf{x}\in\mathbf{X}_{\text{highest}}^{(i)}$ there is an associated ground-truth radiology report $r(\mathbf{x})$. We collect these reports into a set $\mathcal{R}^{(i)}:=\{r(\mathbf{x})|\mathbf{x}\in\mathbf{X}_{\text{highest}}^{(i)}\}$.
We then utilized a pre-trained and frozen language model $f_{\text{descriptor}}:\mathcal{R}\mapsto d$ to generate a description $d^{(i)}$ for a feature $i$ by analyzing the set $\mathcal{R}^{(i)}$ as $d^{(i)}=f_{\text{descriptor}}(\mathcal{R}^{(i)})$. This process yielded a set of feature descriptions $\{d^{(i)}\}_{i=1}^M$, where $M$ is the total number of features learned by the SAE.

For a new scan $\mathbf{x}$ we identified the set of active features $I(\mathbf{x})$ based on a threshold $\tau$ as 
\begin{equation}
    I(\mathbf{x}):=\{i|\mathbf{h}_i(f_{\text{img}}(\mathbf{x}))\cdot\|W^{\text{dec}}\|_2>\tau\}.
\end{equation}
The automated radiology report $R(\mathbf{x})$ is then generated by a pre-trained and frozen LLM from the descriptions of the active features
\begin{equation}
    R(\mathbf{x})=f_{\text{generator}}(\{d^{(i)}|i\in I(\mathbf{x})\}).
\end{equation}

\section{Experiments}
Our overarching hypotheses are that: 1) SAE features capture meaningful visual concepts even in homogeneous datasets (such as is the case for chest radiographs); 2) The visual concepts captured by an SAE can be appropriately described by a pre-trained LLM by use of automated interpretability techniques with paired text data; 3) Natural language descriptions of visual features in the latent space of an SAE can be composed into high-quality radiology reports without explicit use (or training/finetuning) of a VLM for multimodal reasoning. We assessed these hypotheses with our \textbf{automated radiology reporting} experiment (\cref{sec:automating_radiology_reports}). We then conducted a number of \textbf{ablation studies} to investigate the effects of model size, different sparsity constraints, and the inclusion of auxiliary information (\cref{sec:ablation_study}). Next, we performed a case-study for \textbf{counterfactual imagine generation} to evaluate the learned SAE features (\cref{sec:counterfactual_image_main}). Finally, we conducted a \textbf{reader study} with a specialist radiologist to assess the quality of our generated reports (\cref{sec:reader_study}). 
\subsection{Automated radiology reporting}
\label{sec:automating_radiology_reports}
\paragraph{Dataset description} We trained and evaluated all models on the MIMIC-CXR dataset \citep{johnson2019mimic}, a public dataset of 227,835 radiographic studies for a total of 377,110 chest radiographs and associated written text reports. We linked all images to their DICOM metadata files to retrieve scan orientations. We only considered images for which metadata files exist, and retained only antero-posterior(AP)/postero-anterior(PA) scans for training (these are `head-on' scans, as opposed to lateral ones). These are the default views for the generation of diagnostic reports given the increased clarity and ability to visualize the relevant anatomy comprehensively \citep{hyland2023maira}. We then extracted the `findings' section from each text report. The `findings' are a natural language description of all relevant negative and positive features for a given radiograph. Occasionally, the `findings' section is placed into another section with the sub-heading of `impression'. Datapoints without either a findings or impression section were discarded. We used the recommended train/test split for MIMIC-CXR, resulting in a total of 239,931 training and 3,403 test images. 

\paragraph{Evaluation metrics} We evaluated generated radiology reports using both general Natural Language Generation (NLG) metrics (BLEU-4 \citep{papineni2002bleu}, ROUGE-L \citep{lin2004rouge}, METEOR \citep{banerjee2005meteor}) and radiology-specific metrics (RGER score \citep{delbrouck2022improving}, CheXpert F1 score \citep{irvin2019chexpert}). While lexical metrics assess n-gram overlap and word order, clinical metrics like RGER and CheXpert F1 attempt to evaluate factual completeness and consistency by analyzing entity-relationship graphs and predicting common chest X-ray pathologies, respectively. NLG metrics can be inadequate for assessing radiology reports as they don't account for clinical significance \citep{bannur2024maira}, while radiology-specific metrics often rely on specialized models \citep{yu2023evaluating} or pre-specified findings classes \citep{smit2020chexbert, bannur2024maira}. To address these limitations, the RadFact framework \citet{bannur2024maira} uses LLMs to assess sentence-level factuality through bi-directional entailment verification with reference reports| offering a robust evaluation method without relying on pre-specified error types or specialized models. Additional details are provided in \cref{subsec:app_eval_metrics}. 

\paragraph{Experimental setup} The SAE-Rad framework was trained on class tokens produced by the Rad-DINO vision transformer, using an expansion factor of 64 resulting in a latent dimension of 49,152. The model was optimized using Adam with a learning rate peaking at 5e-5 and a sparsity penalty of 8e-3, trained for 200,000 steps with a batch size of 2048. Claude 3.5 Sonnet was used for automated feature interpretation and report generation, while RadFact evaluation employed Llama3-70B-Instruct. Detailed experimental setup details are given in \cref{subsec:app_sae_rad_exp_setup}. 

\begin{table*}[!ht]
\caption{\textbf{Report generation performance on the official MIMIC-CXR test split.}
BL4 = BLEU-4, RG-L = ROUGE-L, MTR = Meteor. Ma-5 (Macro-F1-5), Ma-14 (Macro-F1-14), Mi-5 (Micro-F1-5), and Mi-14 (Micro-F1-14) represent the clinical CheXbert labeler scores. Bolding represents best performance in the current study or between the upper bound models.} 
\small
\centering
\begin{tabular}{@{} l c c c c c c c c c c @{}}
\toprule
\multirow{2}{*}{Model} & \multicolumn{2}{c}{RadFact $\uparrow$ } & \multicolumn{3}{c}{NLG Metrics $\uparrow$ }  & \multicolumn{5}{c}{Clinical Metrics $\uparrow$} \\
\cmidrule(lr){2-3} \cmidrule(lr){4-6} \cmidrule(lr){7-11}
& \multicolumn{2}{c}{F1} & BL4 & RG-L & MTR & RGER & Ma-5 & Ma-14 & Mi-5 & Mi-14 \\
\midrule
\multicolumn{11}{l}{\textit{\textbf{Current study}}} \\
Baseline & \multicolumn{2}{c}{30.0} & 3.1 & 18.6 & 23.5 & 14.9 & 38.3 & 25.5 & 45.0 & 42.8 \\
CheXagent & \multicolumn{2}{c}{36.9} & \textbf{3.7} & \textbf{21.5} & 21.1 & 18.0 & 31.5 & 22.5 & 38.6 & 38.1  \\
SAE-Rad (\texttimes64) & \multicolumn{2}{c}{\textbf{37.2}} & 1.9 & 17.1 & \textbf{29.1} & \textbf{18.2}  & \textbf{47.2} & \textbf{34.3} & \textbf{54.4} & \textbf{53.2} \\
\midrule
\multicolumn{11}{l}{\textit{\textbf{Upper bound models}}} \\
MAIRA-1 & \multicolumn{2}{c}{47.8} & 14.2 & 28.9 & 33.3 & 29.6 & 47.7 & 38.6 & 56.0 & 55.7 \\
MAIRA-2 & \multicolumn{2}{c}{\textbf{50.4}} & \textbf{23.1} & \textbf{38.4} & \textbf{41.7} & \textbf{39.6} & \textbf{50.4} & \textbf{41.6} & \textbf{59.1} & \textbf{58.1} \\
\bottomrule
\end{tabular}

\label{tab:findings_generation_results}
\end{table*}

\begin{figure}[!ht] 
    \centering
    \includegraphics[width=0.9\linewidth]{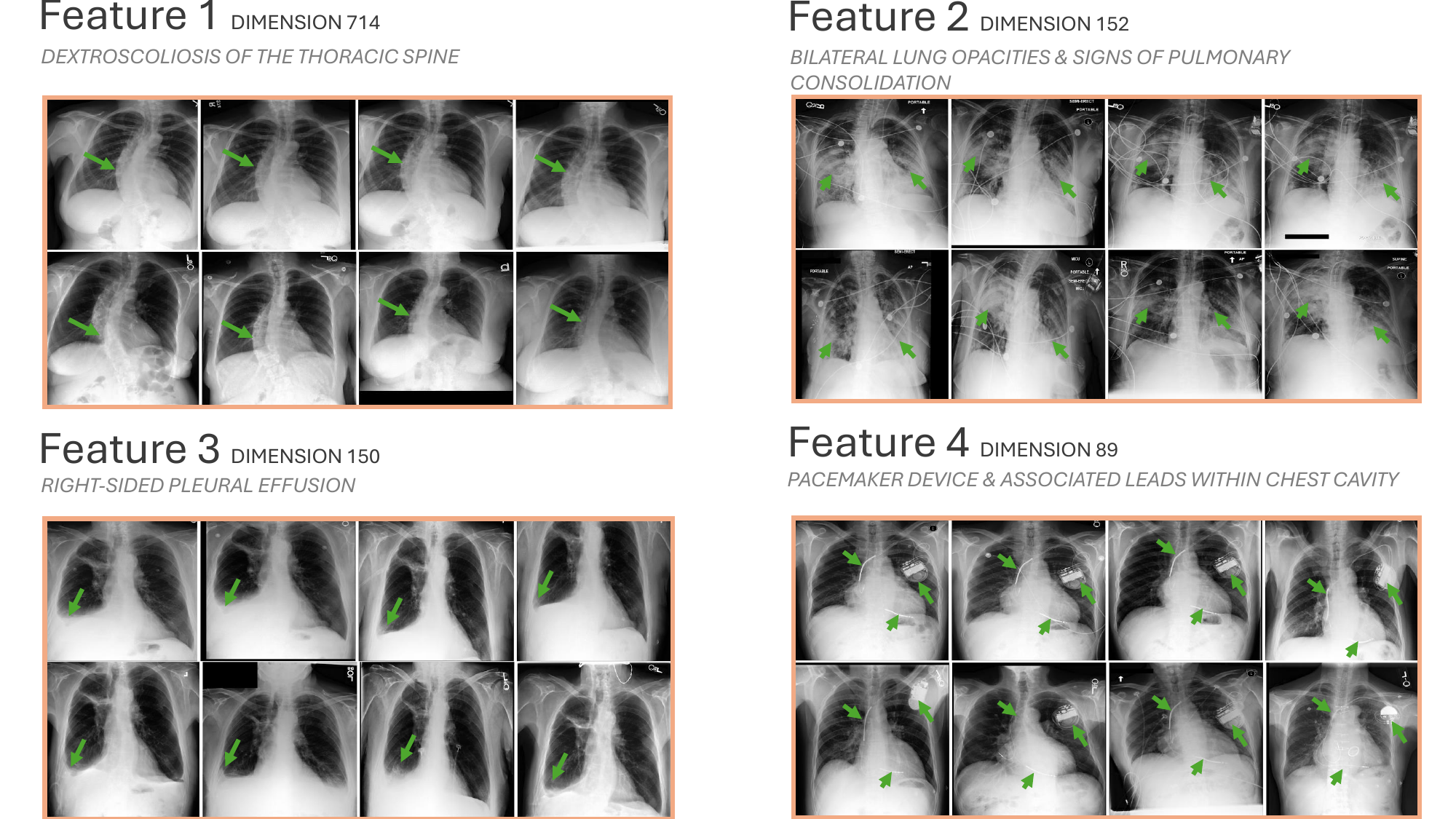}
    \caption{\textbf{SAE-Rad identifies clinically relevant and interpretable features within radiological images.} We illustrate a number of pathological and instrumentation features relevant for producing radiology reports. We add annotations (green arrows) to emphasize the presence of each feature.}
    \label{fig:feature_examples}
\end{figure}
\begin{figure}[!ht] 
    \centering
    \includegraphics[width=0.8\linewidth]{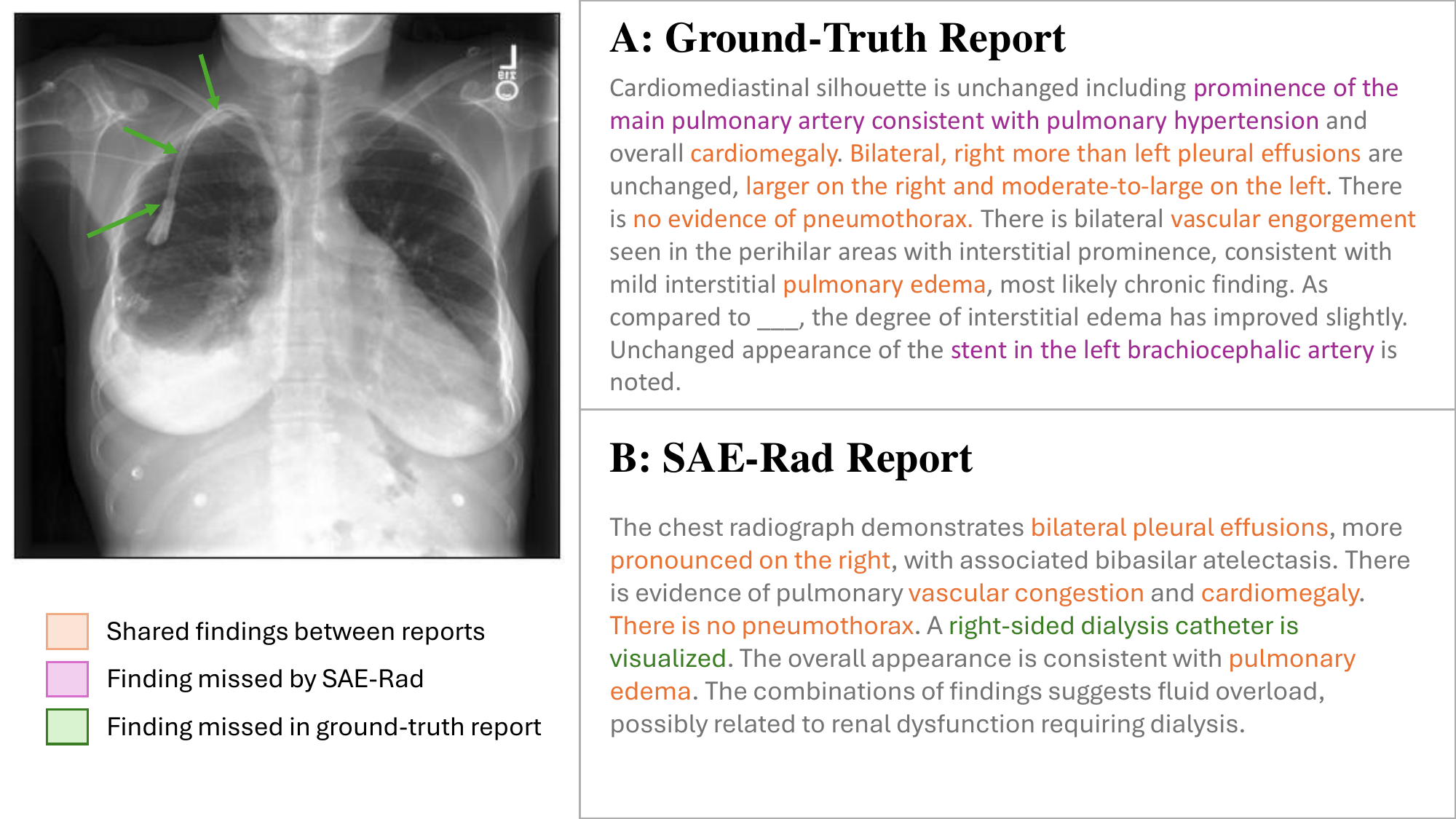}
    \caption{\textbf{SAE-Rad accurately captures features reported by human radiologists and more}. Above, we showcase a side-by-side comparison between a ground-truth radiology report and one generated by SAE-Rad. The model successfully identifies key clinically relevant features. SAE-Rad also identifies additional details, such as a right-sided dialysis catheter, without hallucination (we annotate this feature with green arrows for emphasis). SAE-Rad can also miss features when compared to the reference report.}
    \label{fig:report_example}
\end{figure}
\paragraph{Quantitative evaluation} We compared SAE-Rad to the current state-of-the-art radiology reporting systems. CheXagent \citep{chen2024chexagent} is an instruction-tuned foundation model for CXRs trained on 1.1M scans for question-answering and text-generation tasks. MAIRA-1 \&-2 \citep{hyland2023maira, bannur2024maira} are VLMS based on the LLaVA 1.5 architecture \citep{liu2024visual, liu2024improved}. MAIRA-2 is trained on 510,848 CXRs from four datasets and sets the current state-of-the-art for report generation. The MAIRA systems are not publicly available for result replication, and thus we quote their evaluation values directly as our upper-bound. CheXagent is publicly available, and we therefore performed independent replications for this model for a direct comparison. The `baseline' approach is a na\"{i}ve method of report generation that uses the report of the closest image in the MIMIC train split. Further details of the `baseline' approach can be found in \cref{subsec:baseline_experimental_setup}. As \cref{tab:findings_generation_results} demonstrates, SAE-Rad underperforms on generic NLG metrics such as BLEU-4. This is expected as we do not try to optimize for any specific `writing style' by fine-tuning an LLM on the reference reports from MIMIC-CXR. Conversely, SAE-Rad demonstrates strong performance on radiology-specific metrics which are clinically relevant, outperforming CheXagent by up to 52\% in the CheXpert F1 score (macro-averaged F1-14), and achieving 92.1\% and 89.9\% of the performance of MAIRA-1 and MAIRA-2 on these scores, respectively.    

\paragraph{Qualitative investigation}\label{para:qualitative_invest}
\Cref{fig:feature_examples} illustrates randomly selected monosemantic visual features from SAE-Rad. As can be seen, the SAE learns human-interpretable visual concepts despite the homogeneity and relatively small size of the dataset. These include dextroscoliosis of the spine (\cref{fig:feature_examples}; feature 1), bilateral opacifications (\cref{fig:feature_examples}; feature 2), unilateral pleural effusions (\cref{fig:feature_examples}; feature 3), and the presence of instrumentation -- in this case a pacemaker (\cref{fig:feature_examples}; feature 4). In \cref{fig:report_example}, we illustrate an example `findings' section for a CXR with a number of pathological findings; SAE-Rad is capable of detecting multiple relevant pathologies for a given image. Like other radiology report generation systems, SAE-Rad can miss findings. However, it can also occasionally describe a relevant finding which is otherwise missing from the reference report -- an example relating to the presence of a dialysis catheter is shown in \cref{fig:report_example}.
\subsection{Ablation studies}\label{sec:ablation_study}
We conducted a set of additional experiments which characterize the effects of: 1) Varying the SAE expansion factor; 2) Investigating less sparse (`dense') SAEs by reducing the L1 penalty coefficient; and 3) Evaluating the relative benefits of including auxiliary information including the `indication' section of a report as well as previous reports, where available. The experimental setup for these ablations is described in \cref{sec:appendix_h}.   
\begin{wrapfigure}{r}{0.5\textwidth}
\centering
\scriptsize % This will make the whole table in scriptsize
\setlength{\tabcolsep}{4pt}
\captionsetup{font=small}
\captionof{table}{RadFact performance metrics for different SAE-Rad configurations. /w inds = with indication(s), /w inds + prev.reps = with indications and previous text reports.}
\begin{tabular}{@{}lccc@{}}
\toprule
\textbf{SAE-Rad Configuration} & \textbf{Precision $\uparrow$ } & \textbf{Recall $\uparrow$ } & \textbf{F1 Score $\uparrow$} \\
\midrule
$\times$128                          & 34.83          & 29.91         & 32.18     \\
$\times$64                           & \textbf{35.95} & \textbf{31.95} & \textbf{33.83} \\
$\times$32                           & 31.22          & 27.89         & 29.46     \\
$\times$128\_dense                   & 32.57          & 27.06         & 29.56     \\
$\times$64\_dense                    & 31.15          & 28.46         & 29.74     \\
$\times$32\_dense                    & 32.02          & 28.61         & 30.22     \\
\midrule
$\times$64 /wo inds                  & \textbf{38.78} & 28.22         & 32.67     \\
$\times$64 /w inds                   & 38.45          & 32.42         & 35.18     \\
$\times$64 /w inds + prev. reps      & 37.32          & \textbf{39.83} & \textbf{38.45} \\
\bottomrule
\end{tabular}
\label{tab:combined_metrics}
\end{wrapfigure}

As can be seen in \cref{tab:combined_metrics}, an expansion factor of \texttimes64 produced a higher RadFact F1 score compared with both smaller (\texttimes32) and larger (\texttimes128) expansion factors. In addition, denser SAEs with a larger L0 norm underperformed sparser models. This suggests that concepts useful for radiology report generation are likely to exist in balance between being too coarse (which may cause `feature absorption'| an asymmetric form of feature splitting that can negatively impact the interpretability of an SAE feature \citep{chanind2024absorption}), or too fine-grained, as these features may be more difficult to accurately describe given insufficient amount of descriptive detail in the ground-truth reference reports. \Cref{tab:combined_metrics} demonstrates that the addition of auxiliary information such as the indication, which describes why the patient required the scan in the first instance, can boost the RadFact F1 score, with a large boost to recall. However, in our experiments this caused a small degradation to the precision sub-metric. This finding runs against prior work on the effect of including this section \citep{yu2023evaluating, tu2024towards, nguyen2023pragmatic} and warrants additional investigation in future. We find that adding both previous indications and prior studies has a net positive effect on the quality of generated reports.  
\subsection{Counterfactual image generation for evaluating features}\label{sec:counterfactual_image_main}
\begin{figure}[!ht]
    \centering
    % First subfigure
    \begin{subfigure}{\linewidth}
        \centering
        \includegraphics[width=0.9\linewidth]{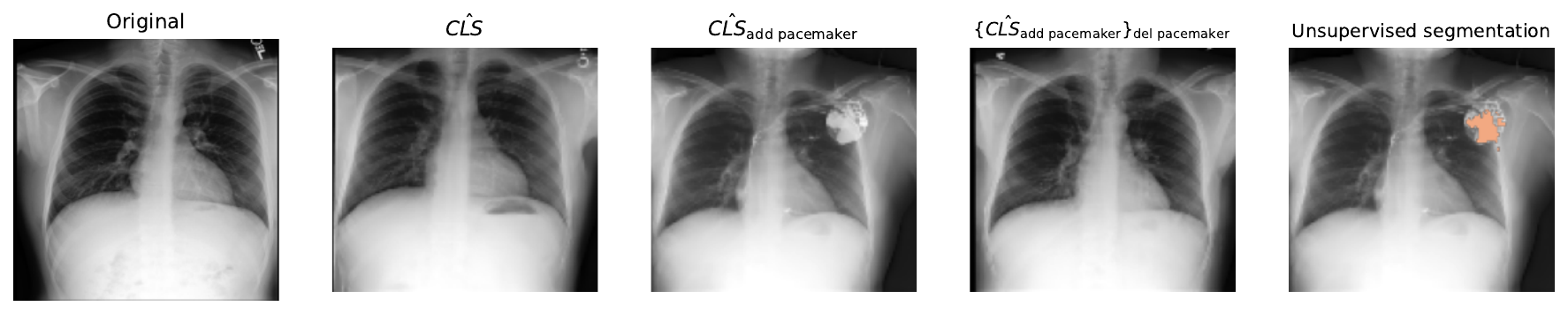}
        \label{fig:pacemaker_intervention}
    \end{subfigure}
    \vspace{1em} % Add vertical space between the subfigures
    % Second subfigure
    \begin{subfigure}{\linewidth}
        \centering
        \includegraphics[width=0.9\linewidth]{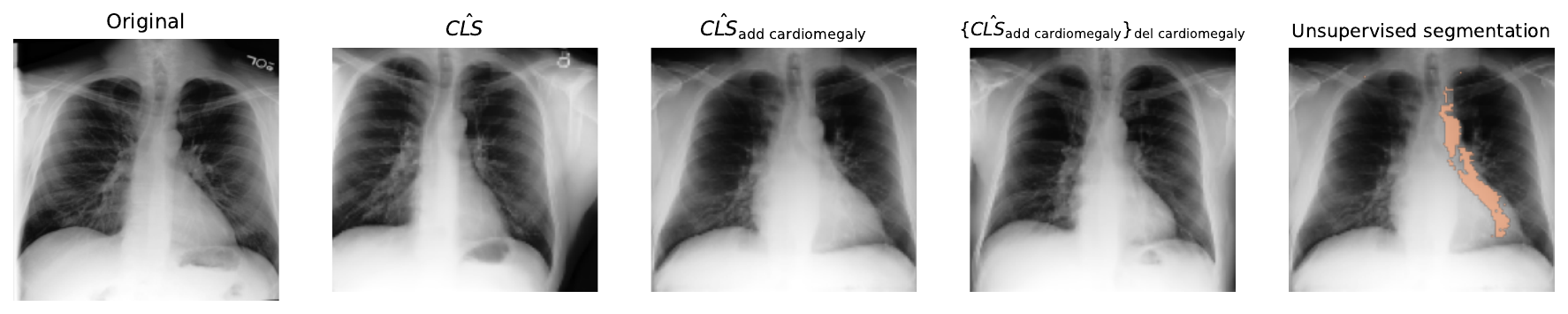}
        \label{fig:cardiomegaly_intervention}
    \end{subfigure}
    \caption{\textbf{SAE-Rad enables counterfactual image generation and unsupervised segmentation with disentangled class tokens.} Row 1 examines a pacemaker, and Row 2 investigates cardiomegaly. Column 1 shows original MIMIC-CXR images, Column 2 shows model reconstructions, and Columns 3 and 4 depict counterfactuals by adding and removing features. The final column demonstrates unsupervised segmentation by comparing counterfactual and original images. Details are in \cref{sec:appendix_d}.} 
    \label{fig:combined_intervention}
\end{figure}

We evaluated the interpretability and validity of our SAE features using an experiment based on \citet{monteiro2023measuring}'s assessment of counterfactual imaging models—focusing on effectiveness, composability, and reversibility. SAE features are interpretable if they correspond to distinct concepts that respond predictably to activation space interventions. We trained a diffusion model conditioned on Rad-DINO class tokens \citep{perez2024rad} to reconstruct MIMIC-CXR radiographs. During inference, we passed a class token through the SAE, intervened on encoder activations, and reconstructed a “counterfactual” token via the decoder, which conditioned the diffusion model to project interventions into imaging space. We tested whether: 1) interventions alter the reconstructed class token accordingly, 2) changes affect only the targeted feature, and 3) features can be “added” or “removed” by manipulating the same activation. \cref{fig:combined_intervention} shows the results for two features (cardiomegaly and pacemaker), demonstrating that our interpretations accurately reflect their impact on model behavior. Further details are given in Appendix \ref{sec:appendix_d}.

\subsection{Reader study}\label{sec:reader_study}
In a reader study with a specialist radiologist evaluating the quality of the automated radiology reports, 165 sentences from 30 reports (SAE-Rad, CheXagent, and a baseline) were analyzed. SAE-Rad had 7\% fewer edits than other models and demonstrated significantly fewer errors with clinical impact, particularly in the ``significant" category, where SAE-Rad had almost half the rate compared to others. This highlights SAE-Rad's potential for radiology report generation in a real clinical scenario. The full study can be found in \cref{sec:appendix_e}.

\vspace{-20px}
\section{Discussion, limitations, and conclusion}
\vspace{-5px}
In this work, we introduced SAE-Rad, a novel framework that leverages sparse autoencoders to automate radiology report generation. Our approach directly decomposes image class tokens from a pre-trained radiology image encoder into human-interpretable features, which are then compiled into comprehensive radiology reports. The experimental results demonstrate that SAE-Rad achieves competitive performance on radiology-specific metrics, outperforming existing models like CheXagent \citep{chen2024chexagent} and approaching the performance of state-of-the-art systems such as MAIRA-2 whilst being trained on a significantly smaller dataset and with a much lower training compute budget; see \cref{sec:appendix_f}. By reverse-engineering the computations of the image encoder, SAE-Rad provides a framework that is verifiably faithful to the underlying model, enhancing transparency and trustworthiness, which are critical considerations in the healthcare setting.

Qualitative analyses confirm that SAE-Rad successfully captures meaningful visual concepts including the presence or absence of pathological features. These interpretable features contribute to generating detailed and accurate radiology reports, as evidenced by strong performance in the clinical evaluation metrics. Our ablation studies indicate that the choice of expansion factor and inclusion of auxiliary information, such as previous reports and indications, can significantly impact the quality of the generated reports, which is broadly in line with the previous literature on this topic REF.

Our approach has limitations. First, SAE-Rad relies on pre-trained (frozen) models for both the image encoder and the LLM in the interpretability pipeline, potentially introducing inherent biases. However, due to the pipeline's modular nature, these biases can be mitigated by replacing either model without retraining the SAE if the LLM is swapped. Additionally, SAE-Rad underperforms on general language metrics like BLEU-4, suggesting that while the generated reports are clinically accurate, they may lack the fluency and stylistic nuances of human-generated reports, particularly those in the MIMIC-CXR dataset. Improving these metrics through style-aware radiology report generation \citep{yan2023style} is a natural avenue for future work. Overall, SAE-Rad presents a novel and effective approach to radiology report generation by leveraging mechanistic interpretability techniques to extract and utilize human-interpretable features from medical images. By providing a verifiably faithful representation of the underlying computations, SAE-Rad contributes to the development of more interpretable AI systems in healthcare. 

\bibliography{paper}
\bibliographystyle{paper}

\newpage
\appendix
\textbf{\LARGE Appendix}
\vskip 8mm

\text{\LARGE{Table of Contents}}
\vskip 4mm
\hrule height .5pt
\vskip 4mm
\begin{itemize}[label={},leftmargin=*]
    \item \textbf{\textcolor{black}{\hyperref[sec:appendix_a]{Appendix A - Reproducibility statement}}} \dotfill \pageref{sec:appendix_a}

    \item \textbf{\textcolor{black}{\hyperref[sec:appendix_b]{Appendix B - Automated radiology report experiment}}} \dotfill \pageref{sec:appendix_b}
    \begin{itemize}[label={},leftmargin=*]
        \item \hyperref[subsec:architecture_comparison]{B.1 SAE architecture comparison} \dotfill \pageref{subsec:architecture_comparison}
        \item \hyperref[subsec:app_eval_metrics]{B.2 Evaluation metrics} \dotfill \pageref{subsec:app_eval_metrics}
        \item \hyperref[subsec:app_sae_rad_exp_setup]{B.3 SAE-Rad additional experimental setup details} \dotfill \pageref{subsec:app_sae_rad_exp_setup}
        \item \hyperref[subsec:prompts_for_sae_features]{B.4 Prompt used to annotate SAE features} \dotfill \pageref{subsec:prompts_for_sae_features}
        \item \hyperref[subsec:prompts_to_compose_findings]{B.5 Prompt used to compose finding sections via SAE features} \dotfill \pageref{subsec:prompts_to_compose_findings}
        \item \hyperref[subsec:baseline_exp_set]{B.6 Baseline experimental setup} \dotfill \pageref{subsec:baseline_exp_set} 
    \end{itemize}
    
    \item \textbf{\textcolor{black}{\hyperref[sec:appendix_c]{Appendix C - Additional qualitative evaluations}}} \dotfill \pageref{sec:appendix_c}
    \begin{itemize}[label={},leftmargin=*]
        \item \hyperref[subsec:more_radiology_reports]{C.1 More radiology report generations} \dotfill \pageref{subsec:more_radiology_reports}
        \item \hyperref[subsec:image_examples_mono_features]{C.2 Image examples of monosemantic features} \dotfill \pageref{subsec:image_examples_mono_features}
    \end{itemize}
    
    \item \textbf{\textcolor{black}{\hyperref[sec:appendix_d]{Appendix D - Counterfactual image generation implementation}}} \dotfill \pageref{sec:appendix_d}
    
    \item \textbf{\textcolor{black}{\hyperref[sec:appendix_e]{Appendix E - Reader study}}} \dotfill \pageref{sec:appendix_e}
    \begin{itemize}[label={},leftmargin=*]
        \item \hyperref[subsec:reader_study_results]{E.1 Results} \dotfill \pageref{subsec:reader_study_results}
    \end{itemize}
    \item \textbf{\textcolor{black}{\hyperref[sec:appendix_f]{Appendix F - Compute efficiency comparisons}}} \dotfill \pageref{sec:appendix_f}
    \item \textbf{\textcolor{black}{\hyperref[sec:appendix_g]{Appendix G - Limits of automated quality assessment for radiology reports}}} \dotfill \pageref{sec:appendix_g}
    \item \textbf{\textcolor{black}{\hyperref[sec:appendix_h]{Appendix H - Ablation experimental setup}}} \dotfill \pageref{sec:appendix_h}

\end{itemize}
\vskip 4mm
\hrule height .5pt
\vskip 10mm
\newpage
\section{Reproducibility Statement}
\label{sec:appendix_a}
We provide details on the experimental setup, hyperparameters, and data preprocessing steps to ensure the reproducibility of our results. We cannot share the MIMIC-CXR dataset, however it can be accessed through application here: \url{https://physionet.org/content/mimic-cxr/2.1.0/}.

\section{Automated radiology report experiment}
\label{sec:appendix_b}
\subsection{SAE architecture comparison}
\label{subsec:architecture_comparison}
In this section we compare the performance of our novel SAE architecture with the architecture proposed by \citet{transformer2024}. The following hyperparameters were used to train both SAEs:

\begin{itemize}
    \item Expansion factor of $\times 64$.
    \item Batch size of 2048.
    \item Learning rate of \num{5e-5}.
    \item Linear warm-up of learning rate for the first 1\% of training.
    \item Linear warm-down of learning rate for the last 20\% of training.
    \item L1 coefficient warmup for the first 5\% of training.
    \item Adam optimizer with no weight decay.
    \item Trained for 200,000 optimization steps.
\end{itemize}

The L1 coefficient was increased for the SAE-Rad architecture in comparison to the \citet{transformer2024} SAE in order to compensate for the additional auxiliary loss term used to train the SAE-Rad architecture. \cref{tab:architecture_comparison} displays the resulting metrics comparing the two SAEs - our proposed SAE attains both a lower L0 and a higher reconstruction accuracy.
\begin{table}[htbp]
\centering
\begin{tabular}{l c c c}
\hline
 & L1 coefficient & L0 $\downarrow$ & Explained Variance (\%) $\uparrow$ \\
\hline
 \citet{transformer2024} SAE & $5.3 \times 10^{-3}$ & 15.7 & 81.3 \\
SAE-Rad & $8.0 \times 10^{-3}$ & \textbf{13.6} & \textbf{84.3} \\
\hline
\end{tabular}
\caption{Comparison of SAE types.}
\label{tab:architecture_comparison}
\end{table}
\subsection{Evaluation metrics}
\label{subsec:app_eval_metrics}
\paragraph{NLG and classical radiology-specific metrics} We evaluated generated radiology reports using both general NLG metrics and radiology-specific metrics. For lexical evaluation, we report BLEU-4 \citep{papineni2002bleu} for 4-gram overlap based on n-gram precision, ROUGE-L \citep{lin2004rouge} for longest common subsequence matching, and METEOR \citep{banerjee2005meteor}, which performs unigram matching using surface forms, stems, and meanings, computing scores based on precision, recall, and fragmentation assessing word order. Whilst widely reported, lexical metrics do not capture factual completeness or consistency \citep{miura2020improving, bannur2024maira} and we therefore also include classical clinical metrics. The RGER score \citep{delbrouck2022improving}, based on the RadGraph model \citep{jain2021radgraph}, evaluates entity-relationship graphs extracted from reports by matching entities and verifying relationships. Additionally, we report the CheXpert F1 score \citep{irvin2019chexpert}, utilizing the CheXbert model \citep{smit2020chexbert} to predict 14 common pathologies in chest X-rays and calculating the harmonic mean of precision and recall between generated and reference texts. Following recommendations \citep{miura2020improving, tu2024towards, hyland2023maira}, we provide micro- and macro-averaged scores over five key observations—atelectasis, cardiomegaly, consolidation, edema, and pleural effusion—as well as the F1-14 score encompassing all observations.

\paragraph{RadFact | A robust evaluation framework for radiology} Natural language generation (NLG) metrics are insufficient to appropriately assess radiology report generations as they do not account for the clinical significance of each sentence, and instead treat all words equally \citep{bannur2024maira}. Conversely, radiology-specific metrics are often based on specialised models such as CheXbert \citep{smit2020chexbert, irvin2019chexpert} or RadGraph \citep{yu2023evaluating, jain2021radgraph, delbrouck2022improving}, which are themselves limited in that rely on specialised models \citep{yu2023evaluating}, pre-specified findings classes \citep{smit2020chexbert, bannur2024maira}, and/or error types \citep{chaves2024training, wang2024llm}. To this end, \citet{bannur2024maira} proposed the RadFact framework which leverages LLMs to assess the factuality of \textit{each sentence} by use of bi-directional entailment verification with the ground-truth reference report. RadFact does not rely on pre-specified error types or radiology-specialized models and is therefore significantly more robust for use to assess the quality of generated radiology reports.

\subsection{SAE-Rad additional experimental setup details}
\label{subsec:app_sae_rad_exp_setup}
In our instantiation of the SAE-Rad framework (described in \cref{sec:sae_rad_sae_arch,sec:sae_rad_full_pipeline}), all bias terms $\mathbf{b}^{\text{enc}}, \mathbf{b}^{\text{gate}},\text{ and } \mathbf{b}^{\text{mag}}$ were initialized to zeros. The elements of the shared encoder weights matrix $W^{\text{gate}}$ were initialized such that all rows point in random directions. The decoder weights $W^{\text{dec}}$ were initialized to $W^{\text{gate} \top}$. For a dataset of size $S$, we trained our SAE on the class tokens $\mathbf{Z}\in\R^{S\times 768}$ produced by the Rad-DINO vision transformer \citep{perez2024rad}, a fine-tuned Dino V2 \citep{oquab2023dinov2} on a large dataset of chest X-rays. We use the model weights released at \url{https://huggingface.co/microsoft/rad-dino}.
The dataset was shuffled and scaled by a constant such that $\mathbb{E}_{\mathbf{z}\in\mathbf{Z}}[\|\mathbf{z}\|_2]=\sqrt{\text{dim}(\mathbf{z})}$, where $\text{dim}(\mathbf{z})=768$. We used an expansion factor of 64 and thus the latent dimension of our SAE is $49,152$. We used the Adam optimization algorithm \citep{kingma2014adam} with $\beta_1=0.9$, $\beta_2=0.999$, and no weight decay. Our learning rate was increased linearly over the first 1\% of training to \num{5e-5} and then decayed linearly to zero over the final 20\% of training. Our sparsity penalty $\lambda$ was linearly increased over the first 5\% of training to \num{8e-3}. We trained our model for $200,000$ steps. We used a batch size of 2048. We did not use ghost gradients or resampling strategies as none of our features were dead at the end of training | in other words, all learnt concept directions were activated at least once when passing the dataset through the SAE. We trained all models on a single NVIDIA A6000 GPU. We used Claude 3.5 Sonnet \citep{anthropic2024claude35} for our automated interpretability pipeline; for a given SAE feature, we retrieved the 10 highest activating images for that feature and passed their ground-truth reference report to Claude with the task of extracting the most consistent information across the reports. The output description was used to label the feature. The full prompts are described in \cref{subsec:prompts_for_sae_features}. 

To generate a report from SAE features, we accrued all activating features and their descriptions, and passed them to Claude for concatenation into a full `findings' paragraph. The full prompts are describe in \cref{subsec:prompts_to_compose_findings}. We evaluate RadFact using Llama3-70B-Instruct.

\subsection{Prompt used to annotate SAE features}
\label{subsec:prompts_for_sae_features} 
In this section we describe the prompt used to instruct Claude 3.5 Sonnet \citep{anthropic2024claude35} to annotate SAE features. Overall, we instruct Claude to identify commonalities between radiological reports corresponding to the 10 maximally activating examples for a given SAE feature. The prompt encourages the model to use chain-of-thought \citep{CoT} reasoning in-order to ensure the quality of the feature annotations. 
\begin{lstlisting}[language=json,firstnumber=1]
You are an expert radiologist specializing in chest radiographs. We're studying neurons in an image neural network, where each neuron detects specific features in chest X-rays. I've identified the radiology images that most strongly activate a particular neuron and will provide you with their associated text radiology reports. Your task is to analyze these reports and determine the common feature that this neuron is detecting.
To arrive at the most accurate and precise explanation of what this neuron is detecting, you must engage in explicit chain of thought reasoning. Begin by thoroughly examining all provided radiology reports, noting any patterns or commonalities. Pay close attention to recurring terminology, described anatomical structures, and consistent pathological findings. Consider how these elements might interrelate to form a singular, distinctive feature that the neuron could be identifying. Evaluate the context of chest radiographs and consider which aspects would be most significant or unique within this imaging modality.
As you progress through your analysis, verbalize your thought process. Explain each step of your reasoning, from initial observations to intermediate conclusions, and finally to your overall assessment. This chain of thought approach will help ensure a comprehensive and well-reasoned final explanation.
After this detailed analytical process, formulate a single, specific explanation of what the neuron is detecting. Your explanation should be as precise and fine-grained as possible, avoiding vague or general statements. Focus on specific features or combinations of features, using 'and' to connect multiple elements if necessary. Avoid using 'or' to list multiple possibilities. Refrain from explaining the pathology itself (e.g., avoid statements like "This feature represents X, which is characterized by..."). Base your explanation solely on the information provided in the reports, without additional medical knowledge that might not be captured by the neuron.
It is crucial that you present your final explanation in the following format:
*This feature represents [your specific, detailed description of what the neuron is detecting].
The asterisk is absolutely essential. Your explanation must begin immediately after the asterisk, without any additional text, numbering, or preamble. The presence of this asterisk is critical for the proper processing of your response.
Below are the radiology reports, listed in order of how strongly they activate the neuron. Use these to inform your analysis and final explanation:
Report number 1: ... 
Report number 2: ...
... 
Report number 10: .
\end{lstlisting} 
Below is an example output from Claude in response to the above prompt.

\begin{lstlisting}[language=json,firstnumber=1]
Thank you for providing these detailed radiology reports. I'll analyze them to determine the common feature that this neuron is detecting. Let me walk you through my thought process:

1. First, I'll review each report, noting recurring elements:
   - Report 1: Mentions cardiomegaly and atherosclerotic calcifications in the thoracic aorta.
   - Report 2: Notes mild cardiomegaly and a moderately tortuous, mildly calcified aorta.
   - Report 3: Describes an enlarged cardiomediastinal silhouette and a calcified, unfolded aorta.
   - Report 4: Mentions moderate enlargement of the cardiac silhouette.
   - Report 5: No specific mention of heart or aorta abnormalities.
   - Report 6: Notes moderate cardiomegaly.
   - Report 7: Mentions moderately enlarged cardiac silhouette.
   - Report 8: Describes moderate cardiomegaly and moderate tortuosity of the thoracic aorta.
   - Report 9: No specific mention of heart or aorta abnormalities.
   - Report 10: Notes an enlarged cardiac silhouette and a tortuous descending thoracic aorta.

2. Analyzing these observations, I notice two recurring themes:
   a) Enlarged cardiac silhouette or cardiomegaly
   b) Abnormalities of the thoracic aorta (tortuosity, calcification, unfolding)

3. Considering the context of chest radiographs, both of these features would be significant and distinctive within this imaging modality.

4. The frequency of these observations is notable:
   - Cardiomegaly or enlarged cardiac silhouette is mentioned in 7 out of 10 reports.
   - Aortic abnormalities are mentioned in 5 out of 10 reports.

5. Given that cardiomegaly is mentioned more frequently, it might be the primary feature the neuron is detecting. However, the aortic abnormalities are often mentioned in conjunction with cardiomegaly.

6. It's possible that the neuron is detecting a combination of these features, as they often co-occur and are related to cardiovascular changes.

7. The neuron might be sensitive to the overall appearance of an enlarged cardiac silhouette combined with changes in the adjacent thoracic aorta, which together create a distinctive pattern on chest radiographs.

Based on this analysis, I believe the most precise and specific explanation for what this neuron is detecting is:

*This feature represents an enlarged cardiac silhouette (cardiomegaly) in conjunction with thoracic aortic abnormalities, particularly tortuosity and calcification.
\end{lstlisting} 

Finally, we use regex to parse the description using the asterisk as a separator.

\subsection{Prompt used to compose finding sections via SAE features}
In this section we describe the prompt used to instruct Claude to compose the finding sections of a radiology report based on: 1) a patient's past radiology reports, if they exist, 2) the features that SAE-Rad has identified are present within the image and  3) the indication for the radiograph.
Each feature is associated to an \textit{importance score} in accordance to the magnitude of the SAE dimension of provenance | the higher the magnitude, the higher the importance score. The scores were normalised to the interval between 0 and 1. Our prompt encourages Claude to focus on features with the highest scores as these were likely to be more significant. See \cref{subsec:more_radiology_reports} for more example outputs following this prompt, alongside their corresponding radiographs. 

\label{subsec:prompts_to_compose_findings}
\begin{lstlisting}[language=json,firstnumber=1]
You are an expert radiologist specializing in chest radiographs. Your task is to write the findings section for a radiology report based on a chest X-ray image. To assist you, I may provide up to three of the patient's past radiology reports, if available. These might contain useful information related to the features of the current scan. I will also give you the indication (reason) for the current X-ray. Additionally, you'll receive text descriptions of features present in the current X-ray image, along with importance scores for each feature. Your primary focus should be on producing the findings section for the latest scan, given the features about that scan. Focus on features with higher importance scores, as these are more prominent in the image and should be emphasized. Assess the current features, and then judge whether it would be appropriate to relate them to information in previous scans, if provided. Do not explicitly mention dates and times from previous reports. Discuss the features present in the X-ray, along with their implications and any deductions you can make. Your response should constitute the 'findings' section of the radiology report, providing a comprehensive analysis of the current X-ray. All of the information is provided below:
<patient_history>
<past_report> 
Report number 1. This report was written 0 years, 2 days, 16 hours and 28 minutes before the current chest x-ray 
INDICATION:  Left-sided pleuritic chest pain ... 
COMPARISON:  Chest radiograph ___ and chest CT ___ ... 
IMPRESSION: Ill-defined patchy opacities in lung bases which may represent ... 
</past_report>
</patient_history> 

<current_chest_x_ray>
<feature 1> 
Feature number 1. Relative importance score 1.0:
This feature represents the absence of pneumothorax, characterized by normal lung appearance at the pleural margins.
</feature 1> 
... 
<feature n> 
</current_chest_x_ray>
Using the information provided, compose the findings section of the radiology report. Be aware that some of the described features may be inaccurate or only loosely related to the actual characteristics present in the X-ray. When faced with conflicting information, rely on the importance scores or a majority consensus to determine which features are most likely correct. In your report, refrain from simply listing the features. Avoid using the word 'feature' entirely in your report. Keep the radiology report brief and to the point. The reason for the current x-ray examination is provided below:

<indication>
Status post CABG.
</indication>

Now write the findings section. This should be a single contiguous paragraph with the findings of the X-ray radiology report. No more than 5 to 6 sentences. Be concise and avoid simply listing the features. Do not respond with any additional text other than the findings. Do not add any concluding statements at the end, only include findings.
\end{lstlisting}

\subsection{Baseline experimental setup}
\label{subsec:baseline_exp_set}
In this section, we provide details on the experimental details of our baseline report generation method. In summary, the baseline approach returns the `closest' report in the train split. Explicitly, we encode a given radiograph using Rad-DINO into a single embedding vector $\mathbf{z} \in \mathbb{R}^{768}$ by selecting the class token. Then, we identify the radiograph $y$ with the closest Rad-DINO class token from the MIMIC-CXR train split $X$ as measured by the Frobenius norm $y = \argmin_{p \in X} || \mathbf{z}(x) - \mathbf{z}(p) ||_{2}$. We then return the original findings section from $y$ as our predicted, baseline report.

\label{subsec:baseline_experimental_setup}

\section{Additional qualitative evaluations}
\label{sec:appendix_c}
\subsection{Additional radiology report generations}

In this section we provide additional qualitative examples of SAE-Rad generated findings sections, highlighting findings jointly described between radiologist generated reports, missing findings by SAE-Rad, and correct findings highlighted by SAE-Rad not present in the original report.
Descriptions and analyses of findings (dis)similarity were produced and reviewed by a clinician.
\label{subsec:more_radiology_reports}
\begin{figure}[!ht]
    \centering 
    \includegraphics[width=1.0\linewidth]{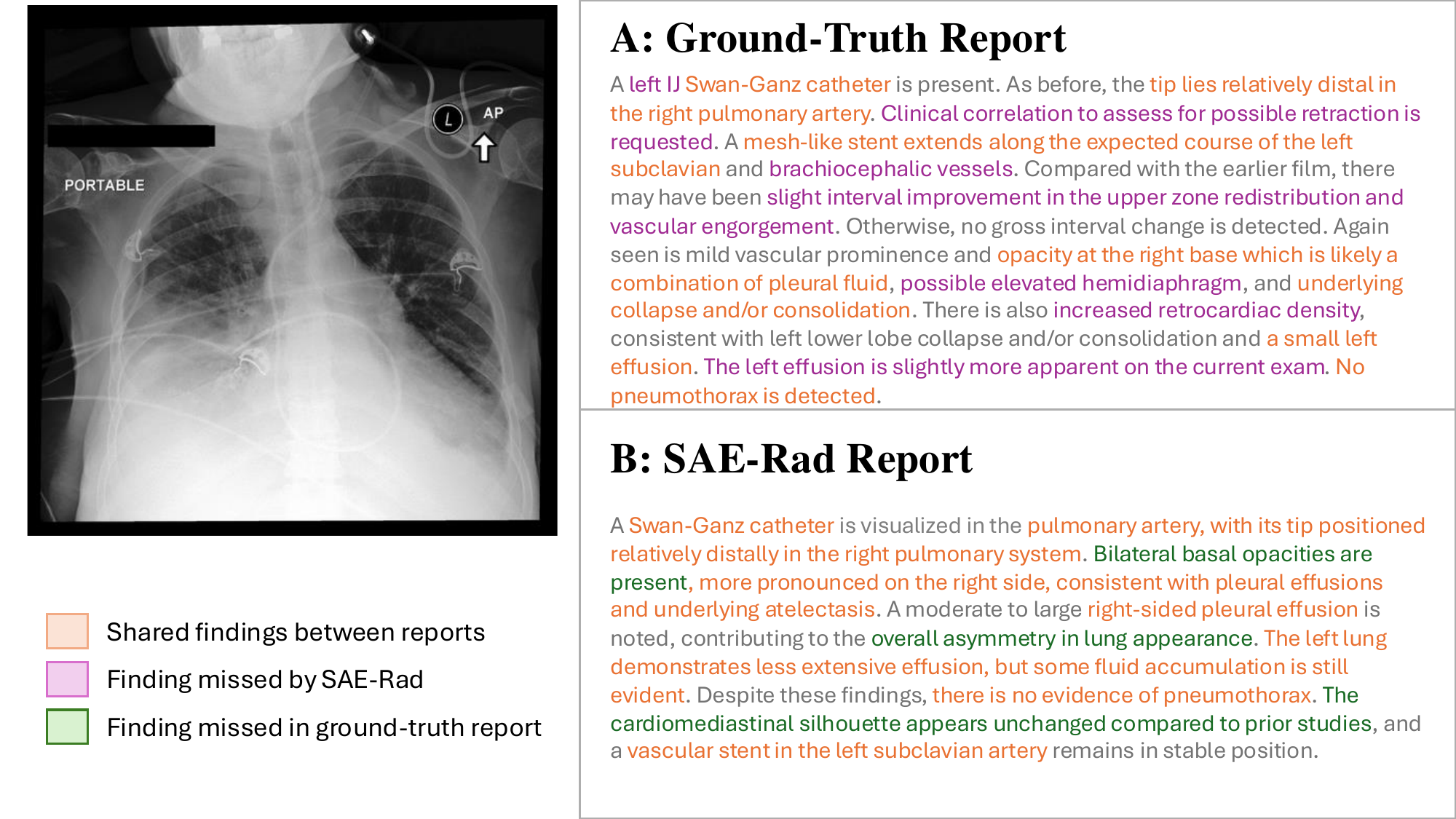}
    \caption{\textbf{Comparison between SAE-Rad Report and the Ground Truth (1)}. Both reports identify a Swan-Ganz catheter terminating distally in the right pulmonary artery, a left subclavian stent, left as well as right (more pronounced) pleural effusion, and no evidence of pneumothorax. SAE-Rad correctly notes bilateral basal opacities, an asymmetry in lung appearance, consistency in the cardiomediastinal silhouette with respect to previous studies, and a moderate to large right pleural effusion. SAE-Rad misses that the stent also extends to the branchiocephalic vessels, and that that the left effusion is slightly more apparent compared to previous exams.}
    \label{1st_app_report} 
\end{figure}
\clearpage 
\begin{figure}[!ht]
    \centering 
    \includegraphics[width=1.0\linewidth]{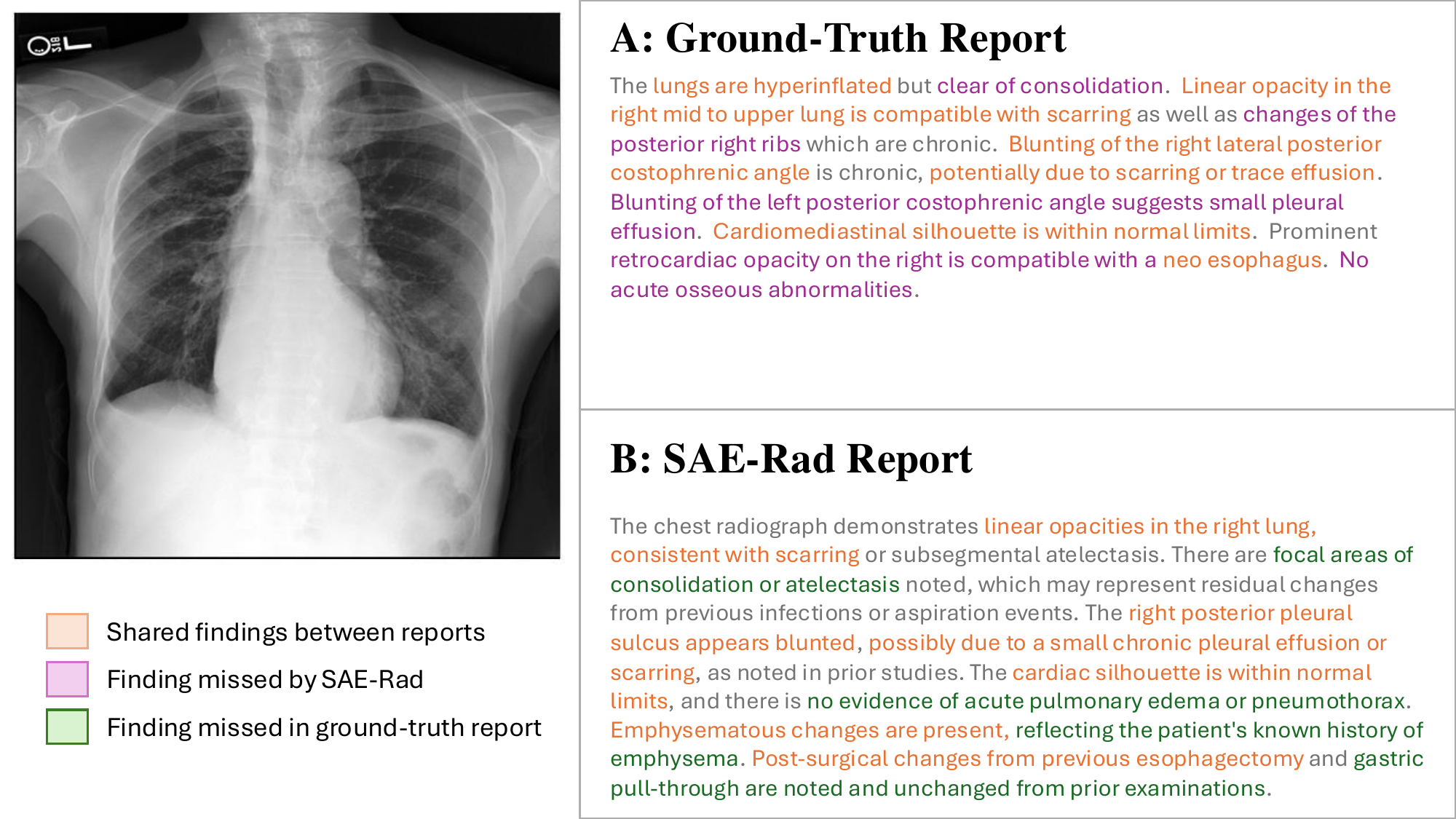}
    \caption{\textbf{Comparison between SAE-Rad Report and the Ground Truth (2)}. Hyperinflation in the lungs, right lung linear opacities, blunting of the right posterior pleural sulcus and the normal limits of the cardiac sihloutee are noted in both reports. SAE-Rad detects the presence of focal areas of consolidation or atelectasis, the patients known history of emphysema, and the gastric pull-through that remains unchanged from prior exams. SAE-Rad misses the changes of the posterior right ribs and the lack of acute osseous abnormalities. }
    \label{2nd_app_report} 
\end{figure}
\begin{figure}[!ht]
    \centering 
    \includegraphics[width=1.0\linewidth]{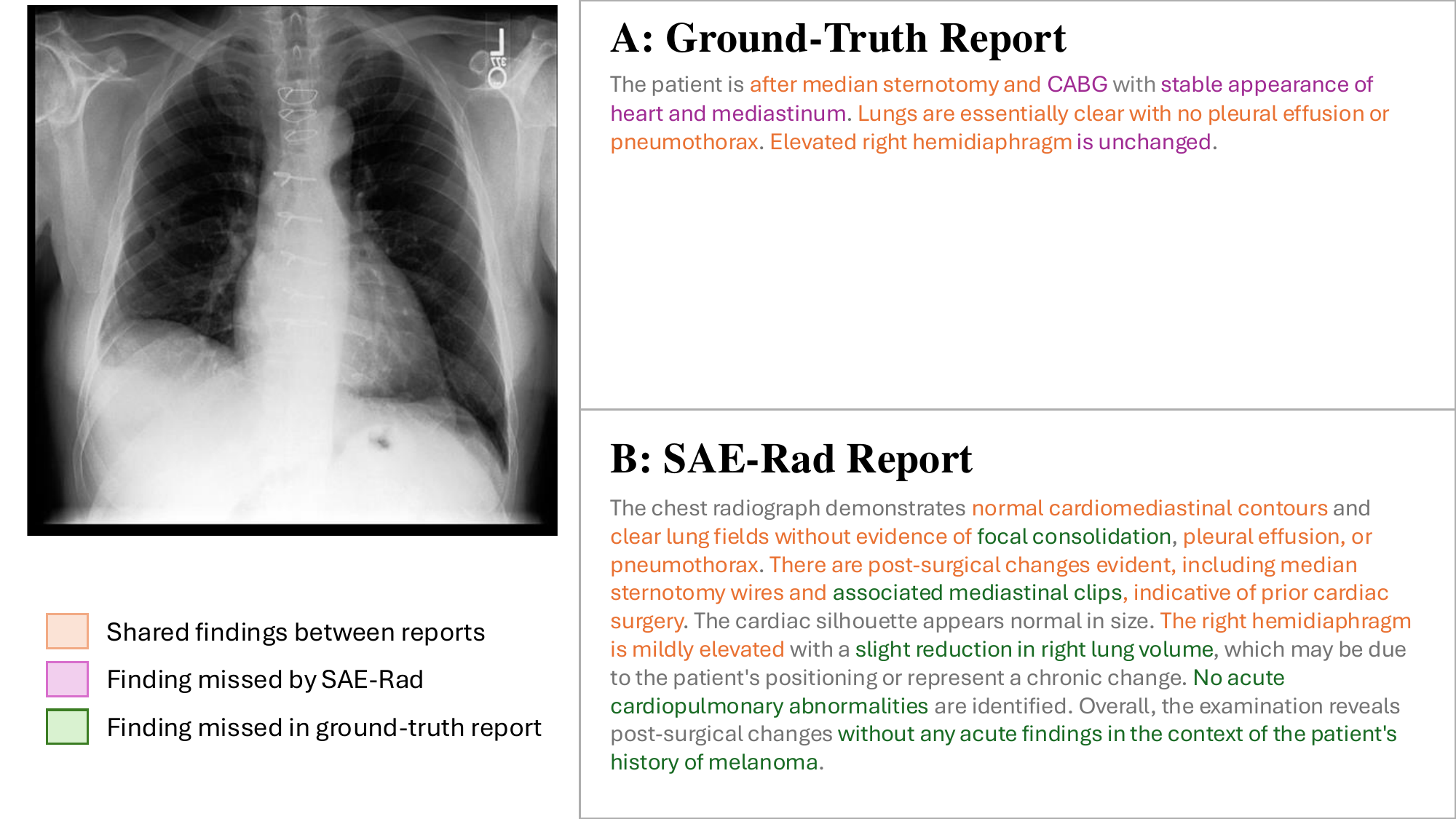}
    \caption{\textbf{Comparison between SAE-Rad Report and the Ground Truth (3)}. Ground-truth and SAE-Rad reports identify the normal catdiomediastinal contours, clear lung fields, as well as the lack of pleural effusion or pneumothorax. SAE-Rad misses that the patient underwent CABG and that their heart, mediastinum and elevated right hemidiaphraghm remain unchanged. SAE-Rad correctly comments on the lack of focal consolidation and cardiopulmonary abnormalities, and notes acute findings with respect to the patients past diagnosis of melanoma. Additionally, SAE-Rad notes the presence of mediastinal clips and a slight reduction in right lung volume.}
    \label{3rd_app_report} 
\end{figure}
\clearpage 
\begin{figure}[!ht]
    \centering 
    \includegraphics[width=1.0\linewidth]{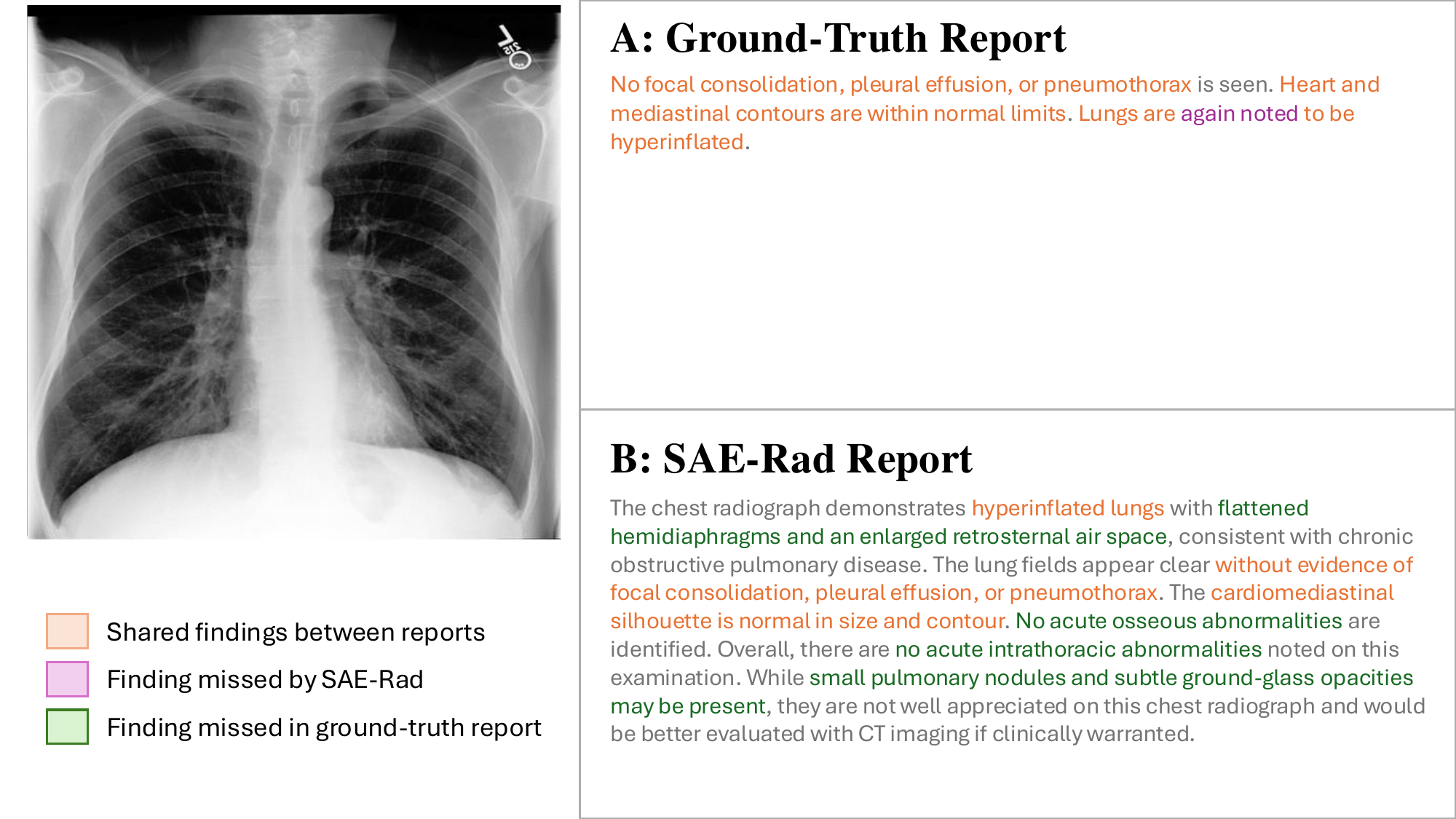}
    \caption{\textbf{Comparison between SAE-Rad Report and the Ground Truth (4)}. In this report, SAE-Rad captures nearly all elements from the ground truth. Both reports note hyper inflation in the lungs, no evidence of focal consolidation, pleural effusion, or pneuomothorax. Additionally, both reports note the normal size and shape of the cardiomediastinal silhoutte. SAE-Rad adds the presence of flattented hemiadiapharghms, an enlarged retrosternal air space, observes small pulmonary nodules, and the lack of acute osseous abnormalities. SAE-Rad misses the comparison to the prior patient history, where lungs were previously noted as hyperinflated.}
    \label{4th_app_report} 
\end{figure}
\begin{figure}[!ht]
    \centering 
    \includegraphics[width=1.0\linewidth]{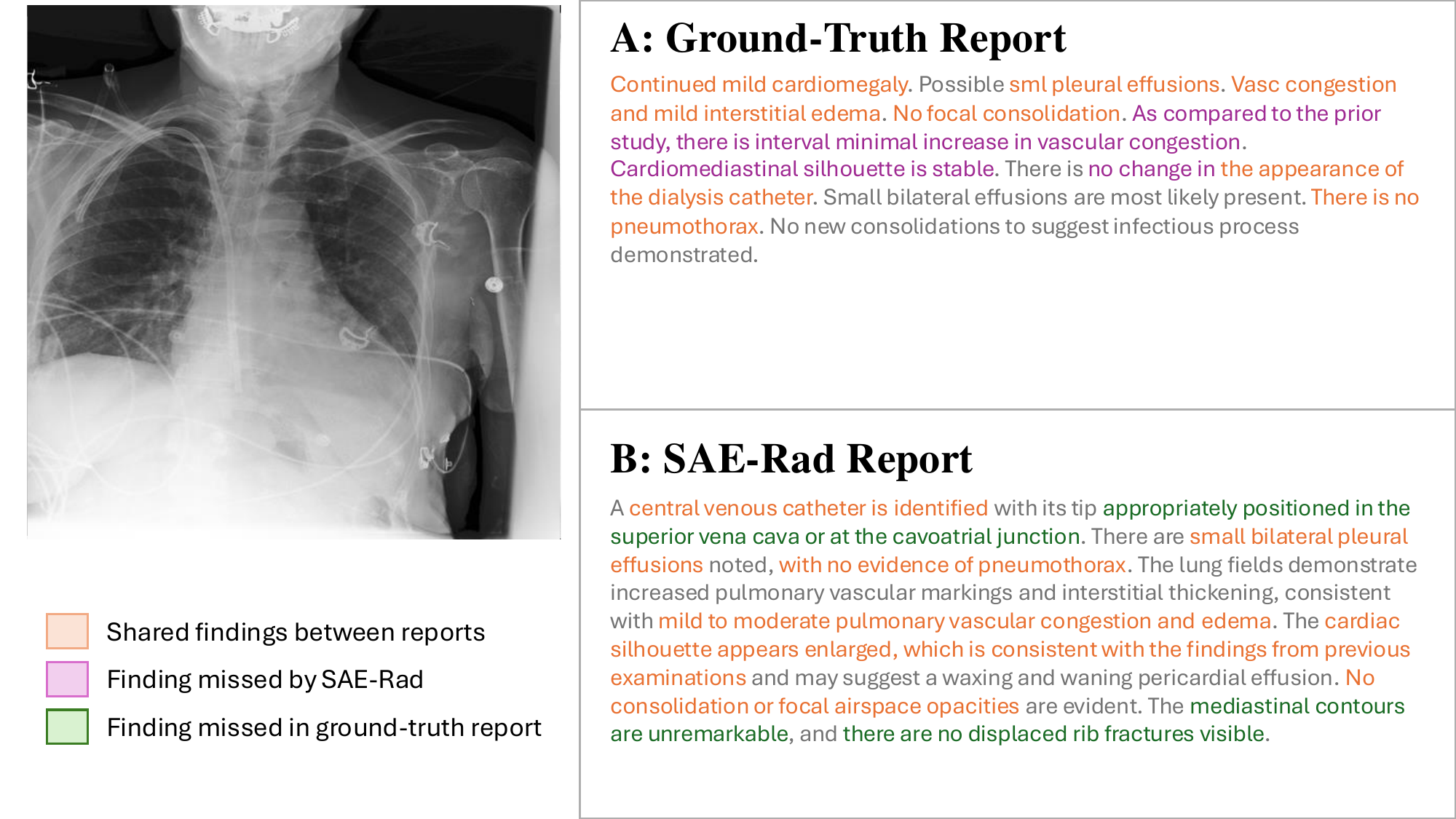}
    \caption{\textbf{Comparison between SAE-Rad Report and the Ground Truth (5)}. The presence of a catheter, as well as small bilateral pleural effusions, an enlarged cardiac silhoutte, and the presences of medium pulmonary vascular congestion with edema is shared across both reports. Both reports also note a lack of pneuomothorax, focal consolidations, and focal airspace opacities. SAE-Rad adds more details on the positioning of the catheter, as well as commenting on the normal shape of the mediastinal contours with no apparent displaced rib fractures. SAE-Rad misses the comparative statements with respect to prior studies in relation to the volume of vascular congestion, the changes in the cardiomediastinal silhouette, and the lack of change of position of the dialysis catheter between studies.}
    \label{5th_app_report} 
\end{figure}
\clearpage 
\begin{figure}[!ht]
    \centering 
    \includegraphics[width=1.0\linewidth]{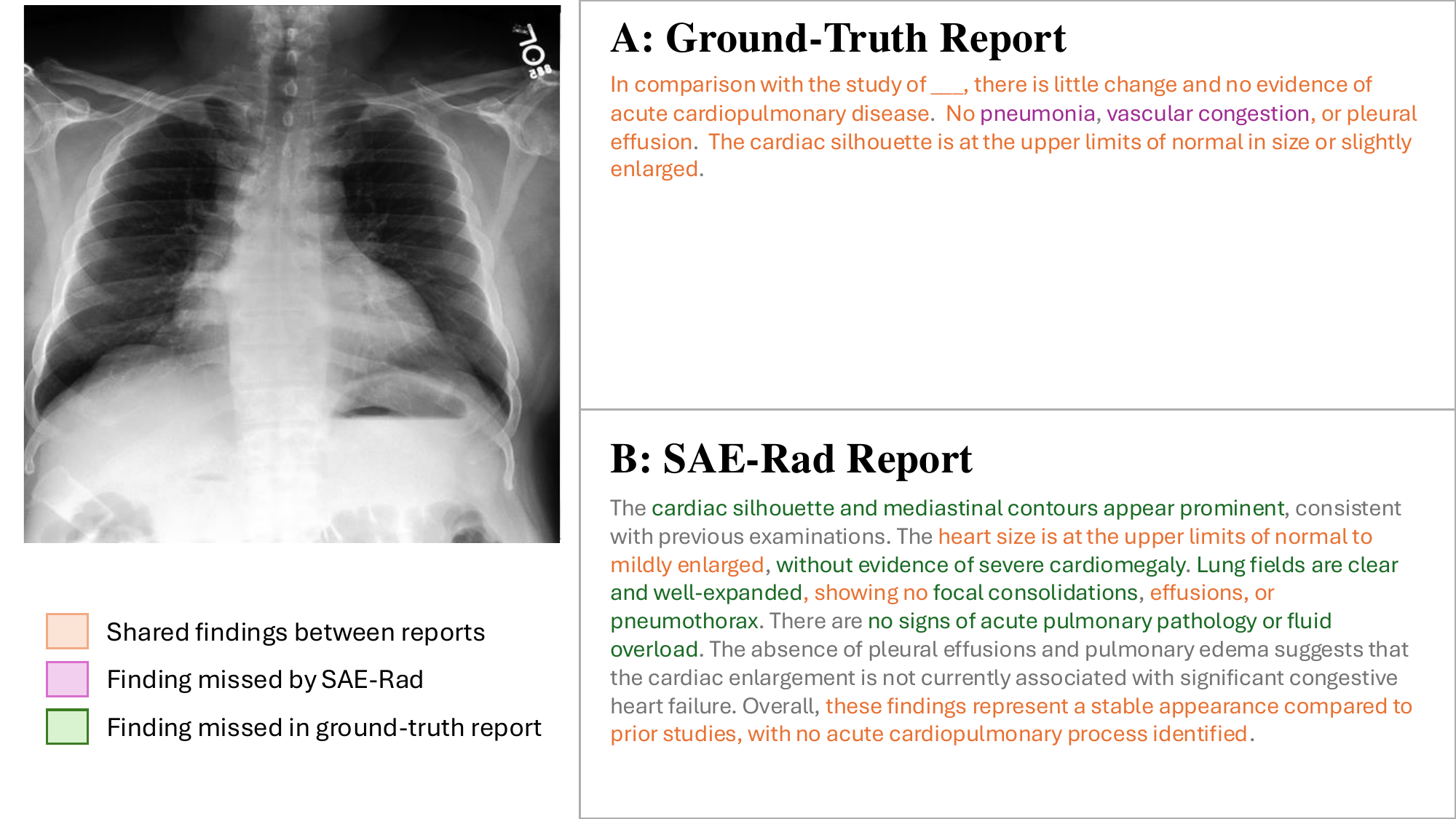}
    \caption{\textbf{Comparison between SAE-Rad Report and the Ground Truth (6)}. In this example, SAE-Rad captures a significant portion of features described in the ground truth report. Both reports note the normal size of the heart, the lack of pleural effusions or acute cardiopulmonary processes, and the stability of appearance with respect to previous exams. SAE-Rad correctly describes the clear appearance of the lungs, as well as the prominence of the mediastinal contours and cardiac silhoutte. Additionally, the lack of severe cardiomegaly, pneumothorax or focal consolidations is noted by SAE-Rad. Comparatively, SAE-Rad fails to note the lack of pneumonia and vascular congestion.}
    \label{6th_app_report} 
\end{figure}
\begin{figure}[!ht]
    \centering 
    \includegraphics[width=1.0\linewidth]{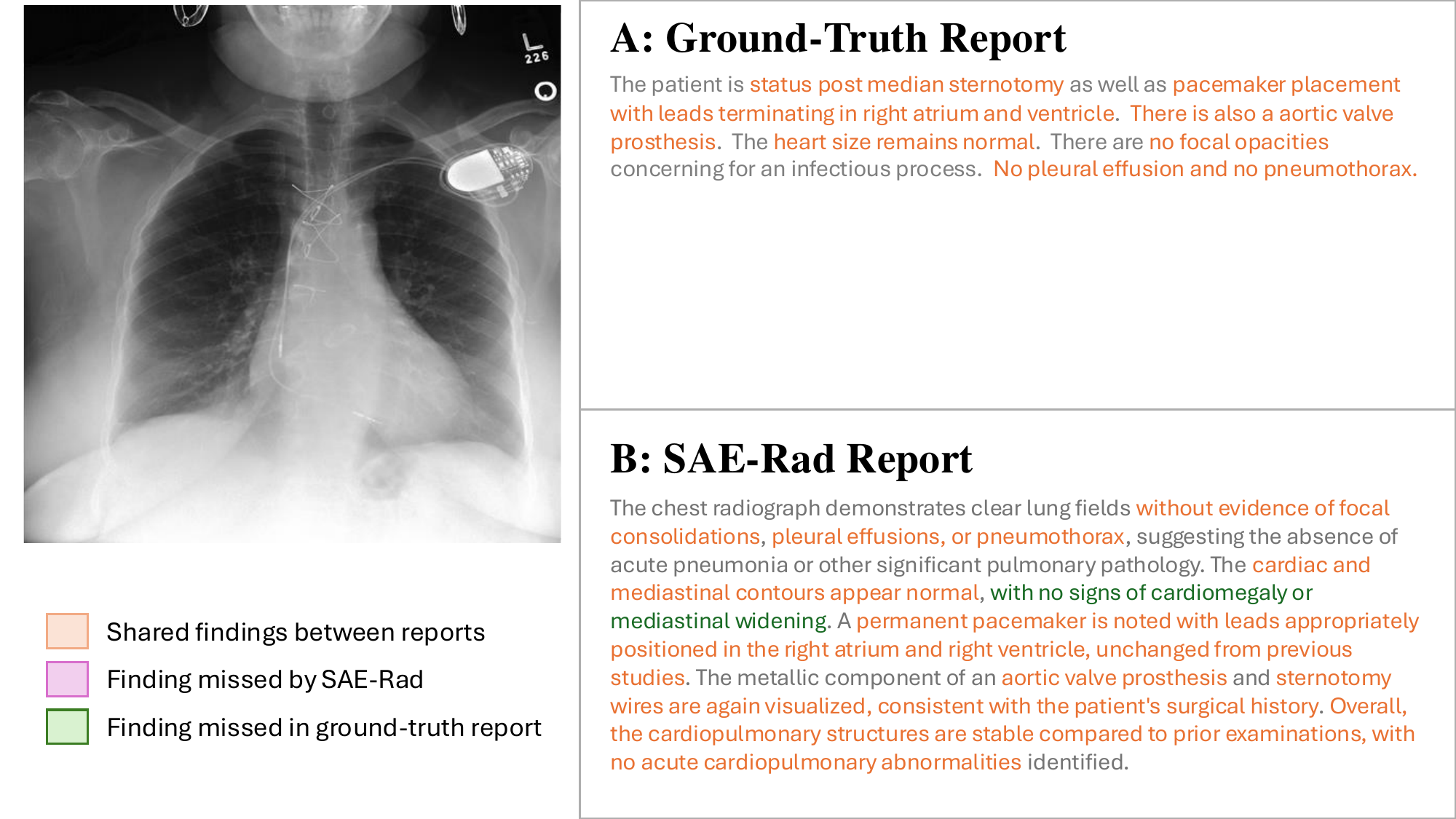}
    \caption{\textbf{Comparison between SAE-Rad Report and the Ground Truth (7)}. SAE-Rad correctly identifies all features in the ground truth. The lack of focal consolidations, pleural effusions and pneumothorax is noted in both reports. Additionally, SAE-Rad describes the normality of the cardiac and mediastinal contours, as well as the stability of the cardiac structures. SAE-Rad correctly identifies reported instrumentation, noting the pacemaker, sternotomy wires in line with the patient's history, as well as the location and presence of an aortic valve prosthesis. SAE-Rad additionally describes the lack of cardiomegaly or mediastinal widening, the stability of the pulmonary system, and the stability of the image with the respect to previous exams.}
    \label{7th_app_report} 
\end{figure}
\clearpage 

\subsection{Image examples of monosemantic features}

\begin{figure}[!ht]
    \centering
    \includegraphics[width=1.0\linewidth]{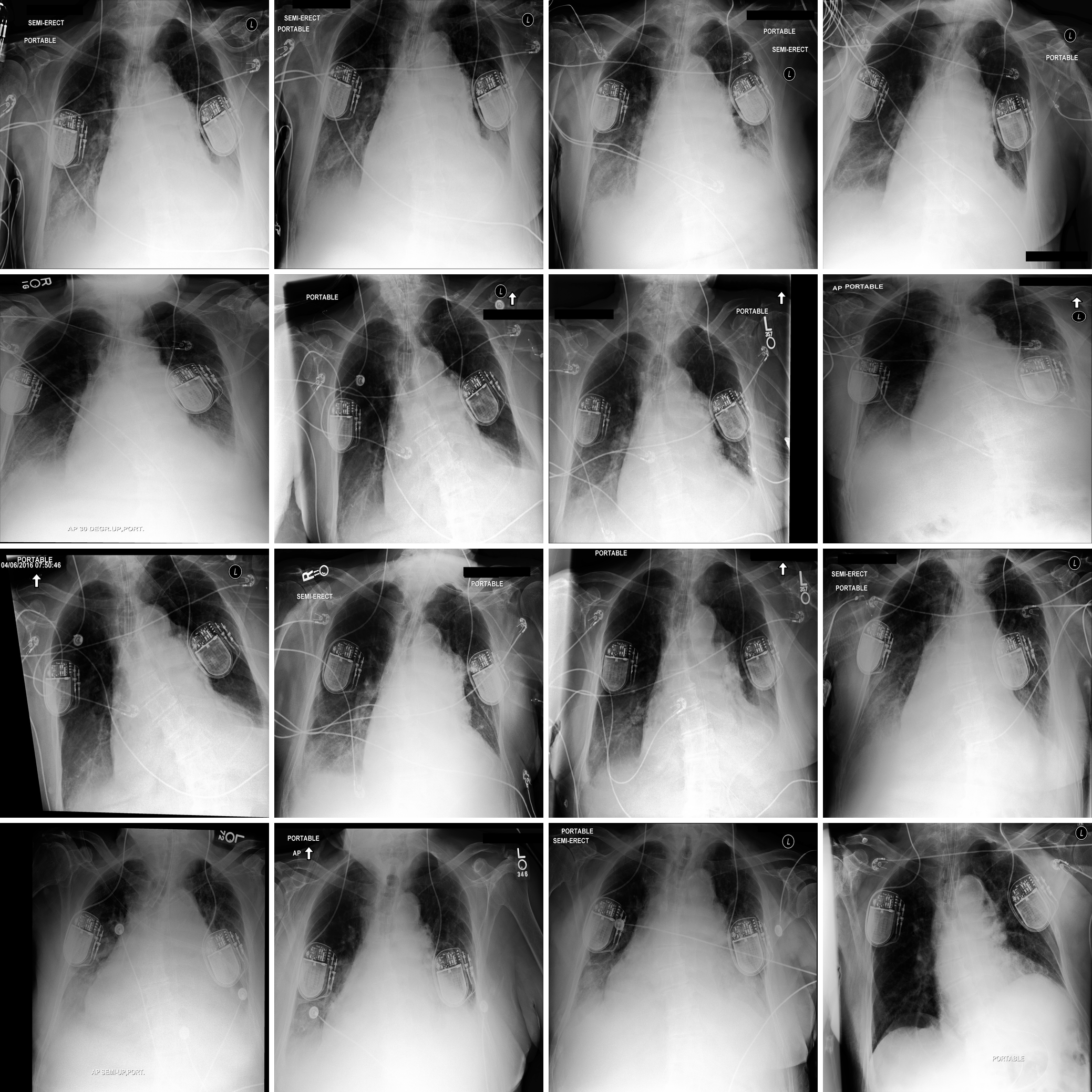}
    \caption{\textbf{Maximally activating images for a feature corresponding to deep brain stimulators which are typically used to treat Parkinson's disease.}}
    \label{image_examples_mono-stimulator}
\end{figure}

\begin{figure}[!ht]
    \centering
    \includegraphics[width=1.0\linewidth]{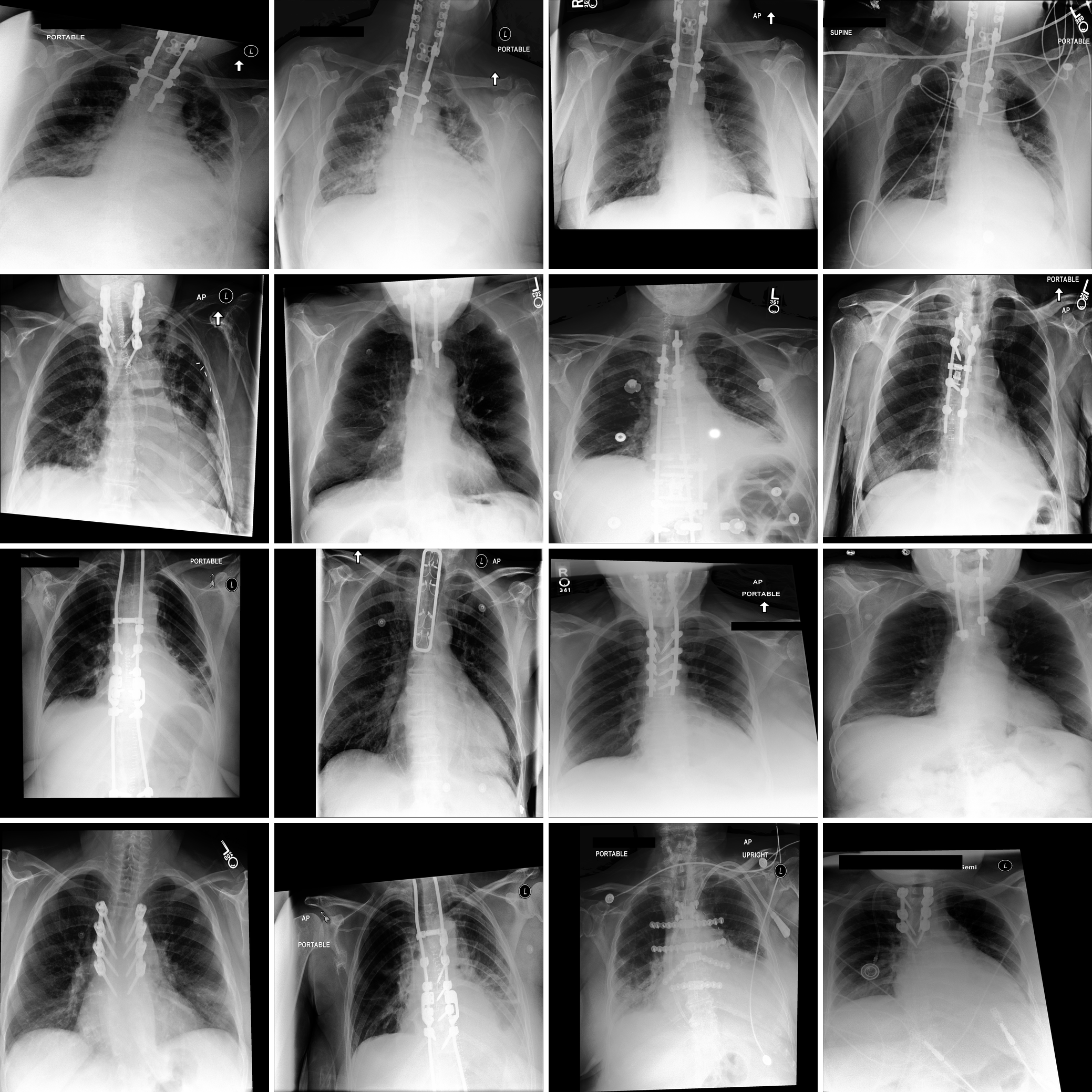}
    \caption{\textbf{Maximally activating images for a feature corresponding to orthopaedic rods and screws.}}
    \label{image_examples_mono-rods}
\end{figure}

\begin{figure}[!ht]
    \centering
    \includegraphics[width=1.0\linewidth]{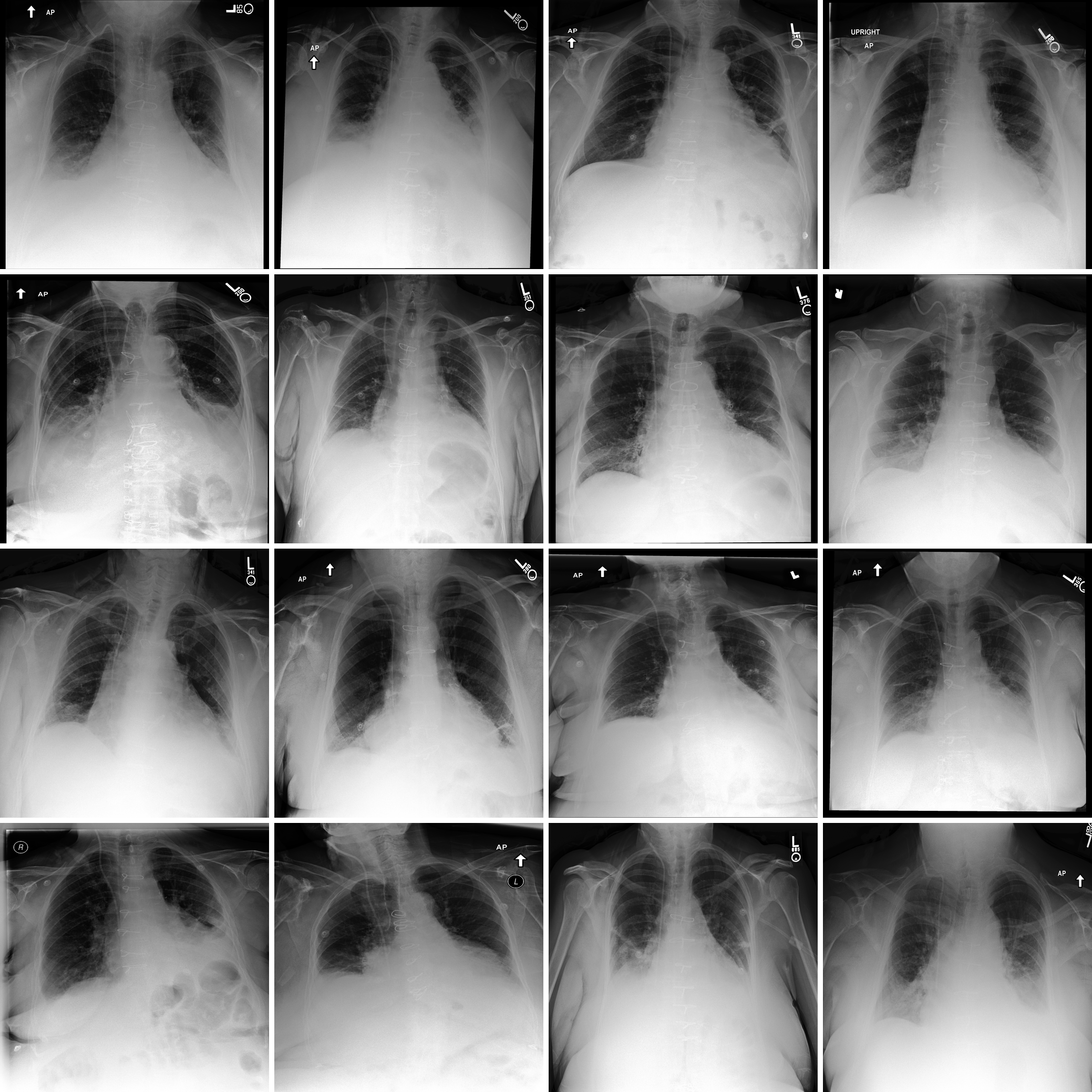}
    \caption{\textbf{Maximally activating images for a feature corresponding to right sided intra-jugular lines in the presence of sternotomy wires.}}
    \label{image_examples_mono-jugular}
\end{figure}

\begin{figure}[!ht]
    \centering
    \includegraphics[width=1.0\linewidth]{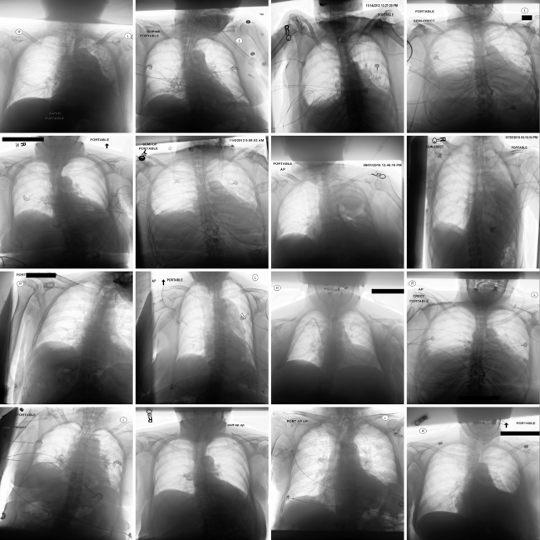}
    \caption{\textbf{Maximally activating images for a feature corresponding to inverted radiographs.}}
    \label{image_examples_mono-inversion}
\end{figure}

\begin{figure}[!ht]
    \centering
    \includegraphics[width=1.0\linewidth]{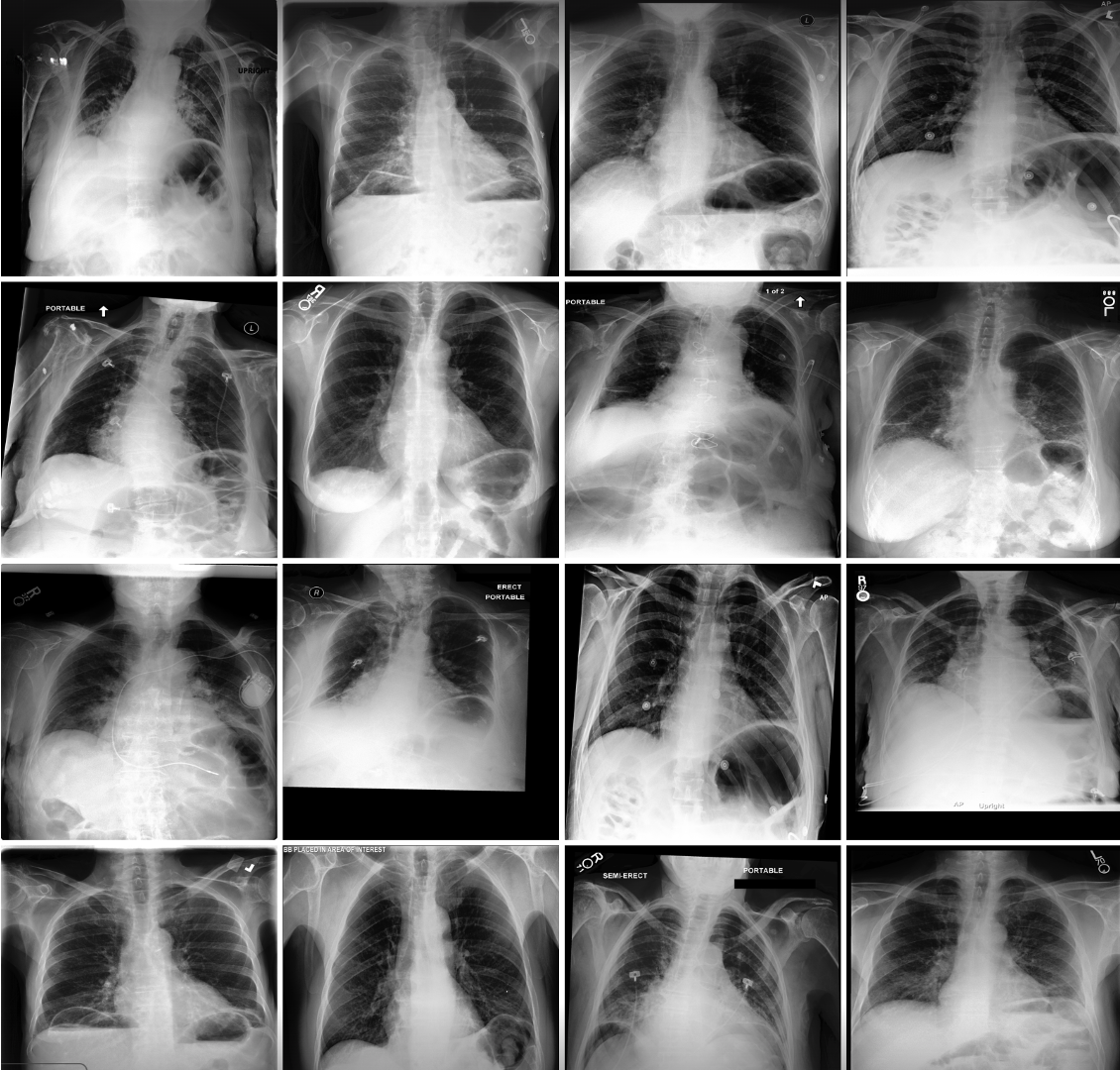}
    \caption{\textbf{Maximally activating images for a feature corresponding to bowel obstruction.}}
    \label{image_examples_bowel_obstruction}
\end{figure}

\begin{figure}[!ht]
    \centering
    \includegraphics[width=1.0\linewidth]{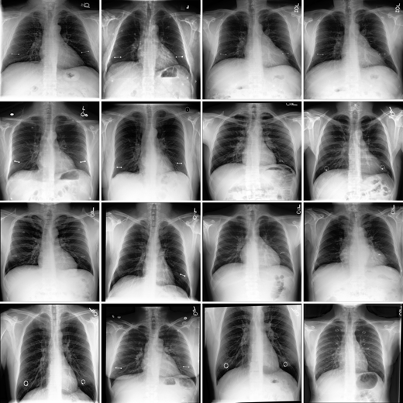}
    \caption{\textbf{Maximally activating images for a feature corresponding to piercings.}}
    \label{image_examples_sae_rad_tps}
\end{figure}

In this section we showcase highest activating images for a number of features, as well as the corresponding feature explanations.
We highlight the variety of features captured by SAE-Rad, from instrumentation features in \cref{image_examples_mono-stimulator}, \cref{image_examples_mono-rods}, \cref{image_examples_mono-jugular}, as well as visual features such as radiograph inversion in \cref{image_examples_mono-inversion}, pathology-related features in \cref{image_examples_bowel_obstruction}, and small details such as piercings in \cref{image_examples_sae_rad_tps}.

\label{subsec:image_examples_mono_features}
\begin{figure}[!ht]
    \centering
    \includegraphics[width=1.0\linewidth]{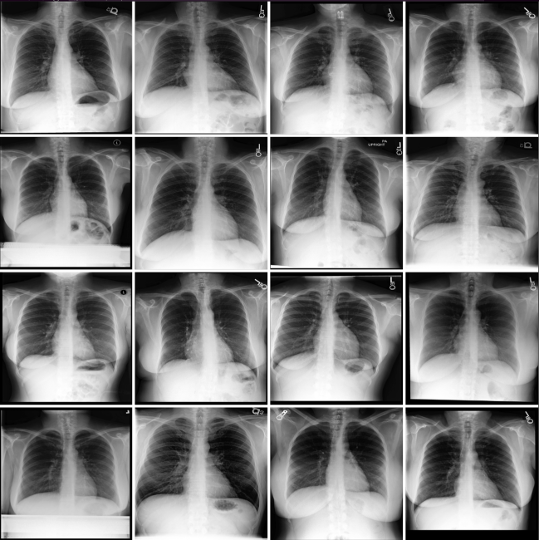}
    \caption{\textbf{Maximally activating images for a feature corresponding to female radiographs with no pathology detected.}}
    \label{image_examples_mono-females}
\end{figure}

\clearpage

\section{Counterfactual image generation implementation} 
\label{sec:appendix_d}

\paragraph{Model Training}
In order to visualise intervention effects, we trained a conditional denoising diffusion probabilistic model (DDPM) \cite{ho2020denoising} to generate chest radiographs given Rad-DINO tokens.
Our DDPM architecture is a three block, 2D UNet, where each block consists of two residual sub-units with SiLU activation and group normalisation.
The downward path consists of a first block of 2D convolutions, followed by spatial self-attention for the second and third blocks.
The upward path consists of two spatial self-attention blocks followed by upsampling for the last block.
The DDPM is conditioned by linearly projecting the CLS token $\mathbf{z} \in \mathbb{R}^{768}$ to a vector $\mathbf{v} \in \mathbb{R}^{1024}$, broadcasting the result to a 2D tensor $\mathbf{j} \in \mathbb{R}^{128 \times 128}$, and concatenating $\mathbf{j}$ channel-wise with a sample $\mathbf{n} \in \mathbb{R}^{128 \times 128}$ as an input to the UNet.
The DDPM was trained for 330 epochs using a 90/10 train/val split of the MIMIC-CXR dataset, with a batch size of 256 and an AdamW optimizer with a learning rate of $0.001$.
We follow standard diffusion model training procedures by randomly sampling timesteps $\{t \in \mathbb{Z} | 0 \leq t \leq 1000 \}$, noising an image $\mathbf{x}_0 \subset \mathbf{X}$ for $t$ steps to obtain noised image $\mathbf{x}_t = \sqrt{\alpha_t} \mathbf{x}_0 + \sqrt{1 - \alpha_t} \epsilon$, where $\alpha_t$ is a scalar controlling the variance of $\mathbf{x}_t$ as a function of $t$, and $\epsilon \sim \mathcal{N}(\mathbf{0}, \mathbf{I})$. 
The objective function can be defined over the mean squared error between diffusion model $\epsilon_{\theta}$ outputs conditioned on class tokens $\mathbf{z}$:
\begin{equation}
\mathcal{L} = \arg \min_{\theta}  \mathbb{E}_{\mathbf{x}_{0} \sim q(\mathbf{x}_{0}), t \sim U(0,1000),  \epsilon \sim \mathcal{N}(0, \mathbf{I})} \left[\|\epsilon^{(t)}_{\theta}(\sqrt{\alpha_{t}}\mathbf{x}_{0} + \sqrt{1 - \alpha_{t}}, \mathbf{z}) - \epsilon\|^{2}_{2} \right] 
\end{equation}
Training was implemented using PyTorch \citep{PyTorch} and the \texttt{diffusers} \citep{diffusers} package on a single LambdaLabs 8xNVIDIA H100 cluster.

\paragraph{Intervention Experiments}
We define a qualitative experiment to assess the visual interpretability of SAE-Rad features by evaluating whether activating or deactivating particular SAE features results in the expected visual changes. 
These inteventions can be defined using a constant reassignment operator
\begin{equation}\label{eqn:do_operator}
    \texttt{do}(i, \beta \ ; \mathbf{h(x)}) : \mathbf{h(x)} \mapsto \mathbf{\tilde{h}(x)}_{i, \beta}
\end{equation}
which sets a particular feature in the SAE activation $\mathbf{h(x)}$ at index $i$ to a desired constant value $\beta$.
Following \cref{eqn:SAE_RAD_eqns}, a given image token $\mathbf{z}$ may be reconstructed into an intervention token $\mathbf{\tilde{z}}_{i, \beta}$ by decoding it's intervened-upon activation $\mathbf{\tilde{h}(x)}_{i, \beta}$

\begin{equation}
    \tilde{\mathbf{z}}_{i, \beta}(\mathbf{h}(\mathbf{z})):=W^{\text{dec}}\texttt{do}(i, \beta; \mathbf{h}(\mathbf{z}))+\mathbf{b}^{\text{dec}}.
\end{equation}

Given a feature of interest $i$, intervention quantity $\beta$, and token $\mathbf{z}$, we can compute a single ``counterfactual" image $\tilde{I}_{i, \beta}$ by reverse denoising $ \tilde{\mathbf{z}}_{i, \beta} + \delta$ using $\hat{\epsilon}_{\theta}$, where $\delta = \mathbf{\hat{z}} - \mathbf{z}$ accounts for the reconstruction error inherent to the SAE between the original token $\mathbf{z}$ and reconstructed $\hat{\mathbf{z}}$. 

Using this framework, we assessed the counterfactual cyclic consistency for a given feature by performing two sequential interventions: we generated $\tilde{I}_{i, \beta}$ by \textit{activating} feature $i$ by $\beta$ for a token $\mathbf{z}$, and subsequently reverted the effects of the intervention by \textit{deactivating} $\beta$ at $i$ in $\tilde{\mathbf{z}}_{i, \beta}$ to generate  $\tilde{I}_{i, 0}$.
Intuitively, $|\tilde{I}_{i, 0} - \tilde{I}_{i, \beta}|$ should represent the visual concepts of interest.
In our experiments, we empirically set $\beta=15$, and select features 311 (pacemaker feature) and 162 (cardiomegaly) our features of interest.
Features were empirically identified by reviewing highest activating feature subsets for our final SAEs.
Results are showcased in \cref{fig:combined_intervention}.

\section{Reader study}
\label{sec:appendix_e}

\begin{figure}[!ht] 
    \centering
    \includegraphics[width=\linewidth]{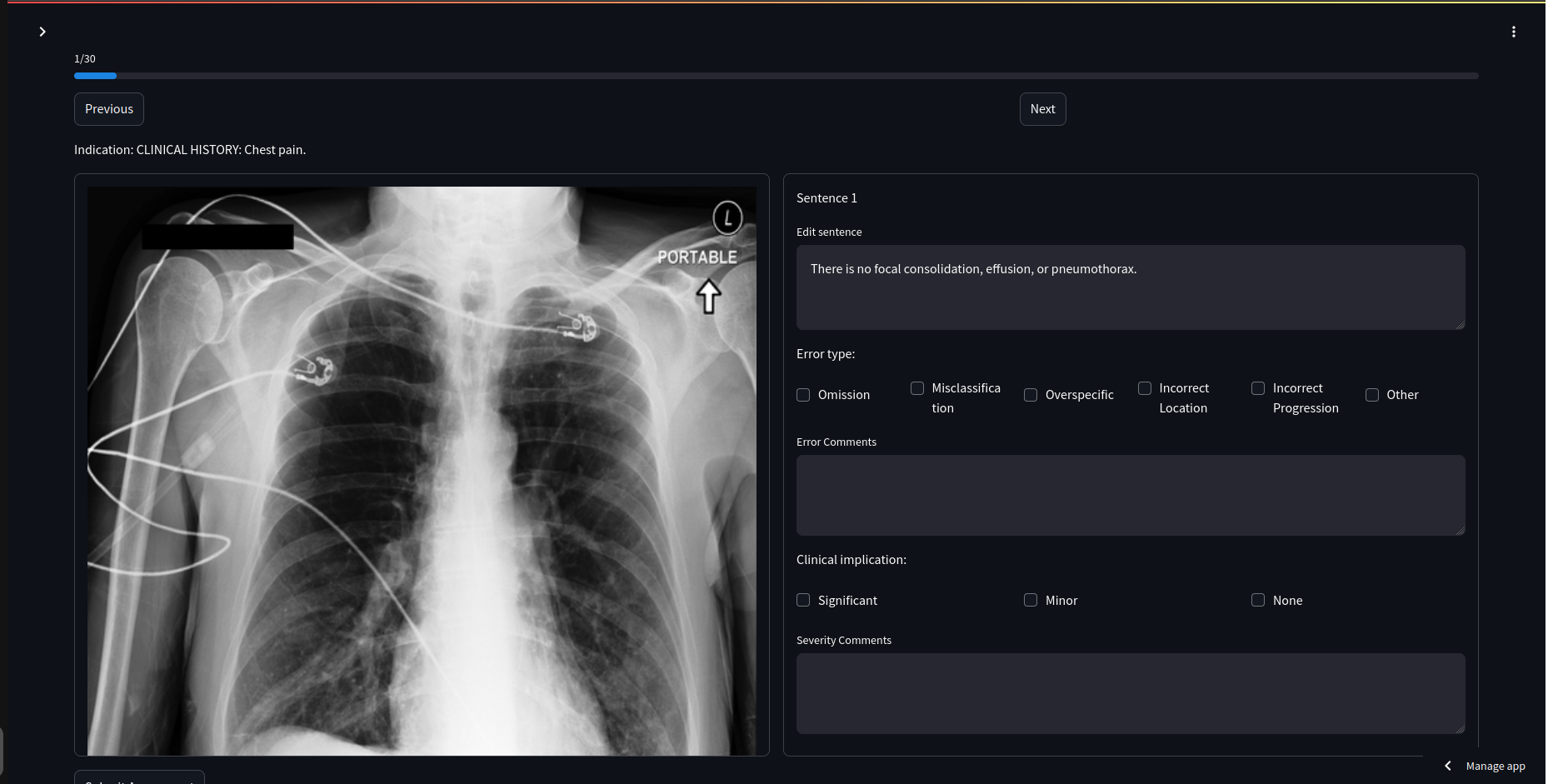}
    \caption{\textbf{Custom UI for radiograph evaluation}.}
    \label{fig:eval_ui}
\end{figure}

In addition to qualitative analyses performed in \cref{para:qualitative_invest}, we performed a reader study to assess radiologist opinions of generated reports compared to currently reported methods.
We followed the reader study design from MAIRA-2 \cite{bannur2024maira} - in summary, we presented a radiologist with a radiograph, the indication for the exam, and the generated report, and subsequently tasked the radiologist with assessing reports sentence by sentence, where the radiologist could edit, delete or maintain the sentences.
Where changes were made, we additionally tasked the radiologist ti categorise errors as one or multiple of ``omission", ``misclassification", ``overspecific", ``incorrect location", and ``other".
We omitted the category ``incorrect progression", as we do not include past radiographs for consideration as part of our reader assessment.
Additionally, unlike the MAIRA-2 reader study, errors could be classified as multiple error categories, which allowed for increased granularity in labelling.
Finally, errors were categorised according to their clinical implications - errors could be ``significant", ``minor", or ``none".
The custom user interface (UI) used for the study is showcased in \cref{fig:eval_ui}.
For full details on error taxonomy, we refer the reader to \cite{bannur2024maira}.

Overall, a radiologist (ST3) reviewed ten unique radiographs with findings section extracted from the ground-truth closest report baseline described in \cref{subsec:baseline_exp_set}, as well as findings generated by CheXagent and SAE-Rad; in total, thirty individual reports were assessed.
The radiologist was blinded to the model that findings originated from during the study.

\subsection{Results}
\label{subsec:reader_study_results}

Overall, a total of 165 sentences were analysed, with an average of 5.5 sentences per report (CheXagent: 6.2, SAE-Rad: 5.9, Baseline: 4.4).
Of 165 sentences, 132 (80\%) sentences were edited or deleted, with 33 (20\%) sentences maintained as is.
Each report, on average, required 4.4 number of edits.
Split by model, CheXagent totalled 52 (83.8\%) edits, the baseline totalled 35 (79.5\%) edits, and SAE-Rad totalled 45 (76.3\%) edits.

Of sentences that required edits, 81 (61\%) sentences overall had no clinical impact, 35 (26.5\%) had a minor impact, and 16 (12.1\%) had a significant impact.
By model, CheXagent had 31 (59.6\%) sentences with edits with no clinical impact, 13 (25.0\%) sentences with minor impact and 8 (15.3\%) sentences with significant clinical impact, the baseline had 19 (54.3\%) sentences with edits with no clinical impact, 11 (31.4\%) sentences with minor impact and 5 (14.2\%) sentences with significant clinical impact, and SAE-Rad had 31 (68.8\%) sentences with edits with no clinical impact, 11 (24.4\%) sentences with minor impact and 3 (6.6\%) sentences with significant clinical impact.

\begin{figure}[!ht] 
    \centering
    \includegraphics[width=\linewidth]{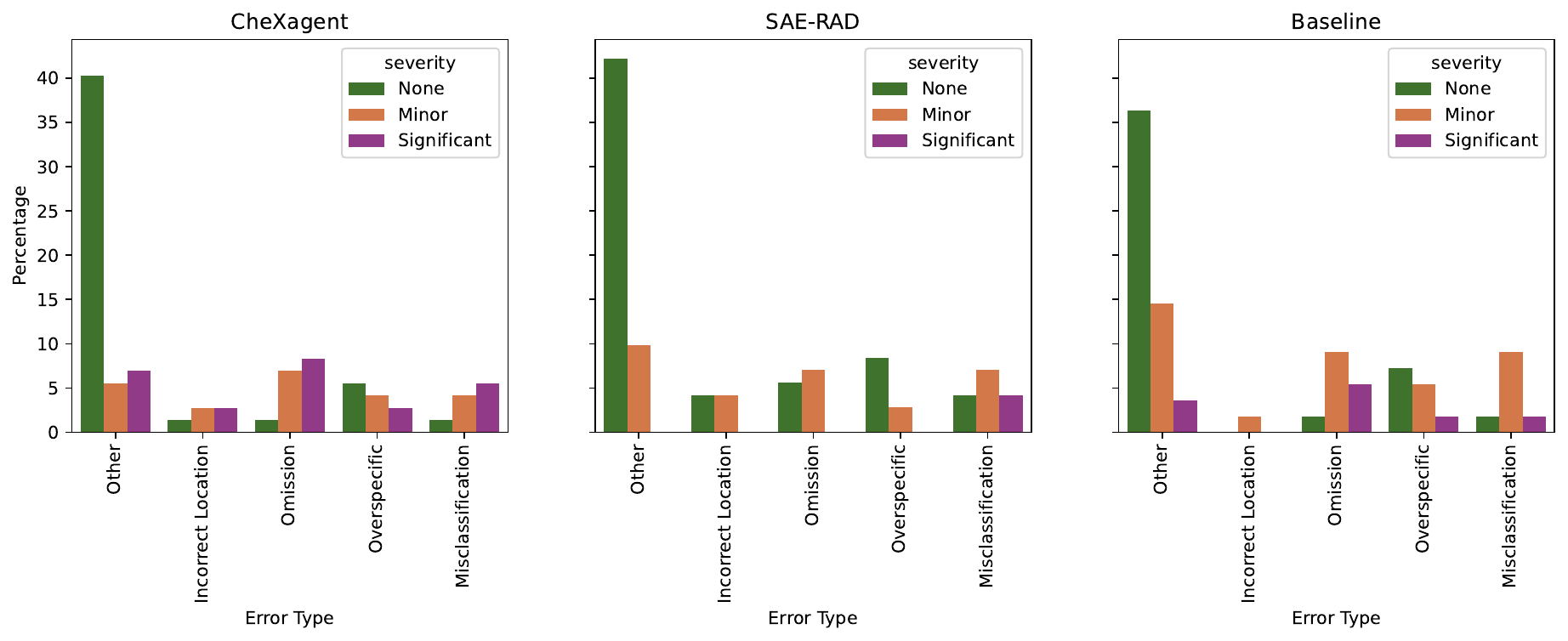}
    \caption{\textbf{Distribution of error types for different models}. Notably, SAE-Rad achieves the lowest number of ``significant" errors in comparison to alternate methods.}
    \label{fig:eval_ui_error_types_model}
\end{figure}

In total, 105 (53.03\%) errors were classified partly as ``other", representing the majority class. 30 (15.15\%) errors were omissions, 26 (13.13\%) errors were mis-classifications, 25 (12.62\%) errors were overspecific, and 12 (6.06\%) errors were incorrect location.
Error type and severity distribution are visualised in \cref{fig:eval_ui_error_types_model}.
Notably, whilst many ``other" type errors were observed, the majority of these errors carried no clinical risk.
In contrast, the ``omission" and ``misclassification" categories convey much higher risk, with a comparatively higher proportion of minor and significant severity categories.
Overall, for sentences with edits, each sentence was labelled with an average 1.46 $\pm$ 0.67 error categories (SAE-Rad: 1.56 $\pm$ 0.62, CheXagent: 1.35 $\pm$ 0.71, baseline: 1.51 $\pm$ 0.62 error categories per edited sentence).

Overall, SAE-Rad requires 7\% fewer edits than comparable methods.
Additionally, significant clinical impact error rates are almost half reports generated using our method compared to other models, whilst maintaining comparable minor and no impact error types.
Similarly to results reported by \citet{bannur2024maira}, the majority of errors have no clinical impact.
Overall, this showcases the potential our proposed method for radiology report generation in a real clinical scenario.

\section{Compute efficiency comparisons} 
\label{sec:appendix_f}
In this section we provide and discuss the compute estimates for training and inference of SAE-Rad in comparison to both MAIRA 2 and CheXagent. In \cref{tab:compute-comparison} we provide quantitative estimates of the compute efficiency - we estimate that SAE-Rad uses approximately 1000x less train compute, 100x less parameters and 4x less data. 
We note that while in this paper we combine the SAE text features into a report using Claude, our method enables the generation of reports using a much smaller LLM, or alternately without an LLM | for example, by concatenating the SAE feature text explanations into a report using only regex. This reduces the inference FLOPs and cost to a negligible quantity. In \cref{tab:compute-comparison}, we have estimated the lower bound inference compute requirements for SAE-Rad using this assumption.
\begin{table}[!ht]
\centering
\begin{tabular}{|l|c|c|c|}
\hline
 & SAE-Rad & cheXagent & MAIRA 2 \\
\hline
Number of unique training images & 239,931 & 1,100,000 & 510,848 \\
\hline
Active training parameters & 76M & 8B & 7B \\
\hline
Estimated training Tera-FLOPs & 60,000 & NA & 100,000,000 \\
\hline
Estimated training cost & \$3.25 & NA & \$4000 \\
\hline
Estimated inference Tera-FLOPs per report & 0.1 & 8 & 7 \\
\hline
Estimated inference cost per report & \$0.00005 & \$0.001 & \$0.001 \\
\hline
\end{tabular}
\caption{Comparison of compute resources for different models.}
\label{tab:compute-comparison}
\end{table}

\section{Limits of automated quality assessment for radiology reports}
\label{sec:appendix_g}
As described in \cref{subsec:app_eval_metrics}, NLG metrics such as BLEU-4 \citep{papineni2002bleu}, ROUGE-L \citep{lin2004rouge}, and METEOR \citep{banerjee2005meteor} do not account for the clinical relevance of the sentences composing a radiology report. Radiology-specific metrics such as the RGER score \citep{delbrouck2022improving} and the CheXpert F1 score \citep{smit2020chexbert} were consequently developed, seeking to improve upon these limitations by using radiology-specialized models as backbones to compute clinically relevant commonalities between generated and reference reports.
However, CheXpert and RGER rely on pre-specified findings classes and error types \citep{bannur2024maira}, which limits their ability to assess the quality of generated reports when elements outside these specified categories are present. RadFact \citep{bannur2024maira} was recently proposed to address this limitation; by relying on the logical inference capabilities of large language models (LLMs), RadFact evaluates the correctness and completeness of generated reports outside of pre-specified classes. 

RadFact introduces two key metrics: RadFact logical precision and RadFact logical recall, henceforth denoted as RadFact precision and recall.
RadFact precision calculates the proportion of generated sentences that are entailed by the ground-truth report, assessing the truthfulness of the model's outputs by penalizing hallucinations. RadFact recall measures the proportion of ground-truth sentences that are entailed by the generated report, evaluating the completeness of the generated report by penalizing omissions.

A limitation of RadFact is that it relies on the assumption that the ground-truth report used as a basis for evaluation is a gold-standard label which exhaustively captures clinically relevant detail.
Therefore, in cases where a predicted report correctly includes findings present in the image but omitted from the ground-truth report, RadFact precision will drop. This means that the metric may not fully reflect the model's ability to generate clinically accurate and comprehensive reports, when the ground-truth reports are incomplete.

We illustrate potential pitfalls of RadFact in \cref{radfact_limits} below.
In this case, the ground-truth report fails to report several clinically relevant details observed in the SAE-Rad report, including, for example, the normal size of the cardiac silhouette, and the increased density in the right upper lung field.
We provide additional examples of this phenomenon in \cref{subsec:more_radiology_reports}.

We note that the level of detail for radiographic reports is likely impacted by multiple factors, such as report indication, patient history, urgency of the request, and care setting of the radiograph.
Given we perform evaluation on MIMIC-CXR |a dataset of radiographs captured at an emergency department \citep{johnson2019mimic}| SAE-Rad is likely to report clinically correct and relevant statements which are not explicitly mentioned by radiologists.
Therefore, reported metrics must be interpreted with care, and may be underestimates of actual clinical factuality and relevance of generated reports. 

\begin{figure}[!ht]
    \centering 
    \includegraphics[width=1.0\linewidth]{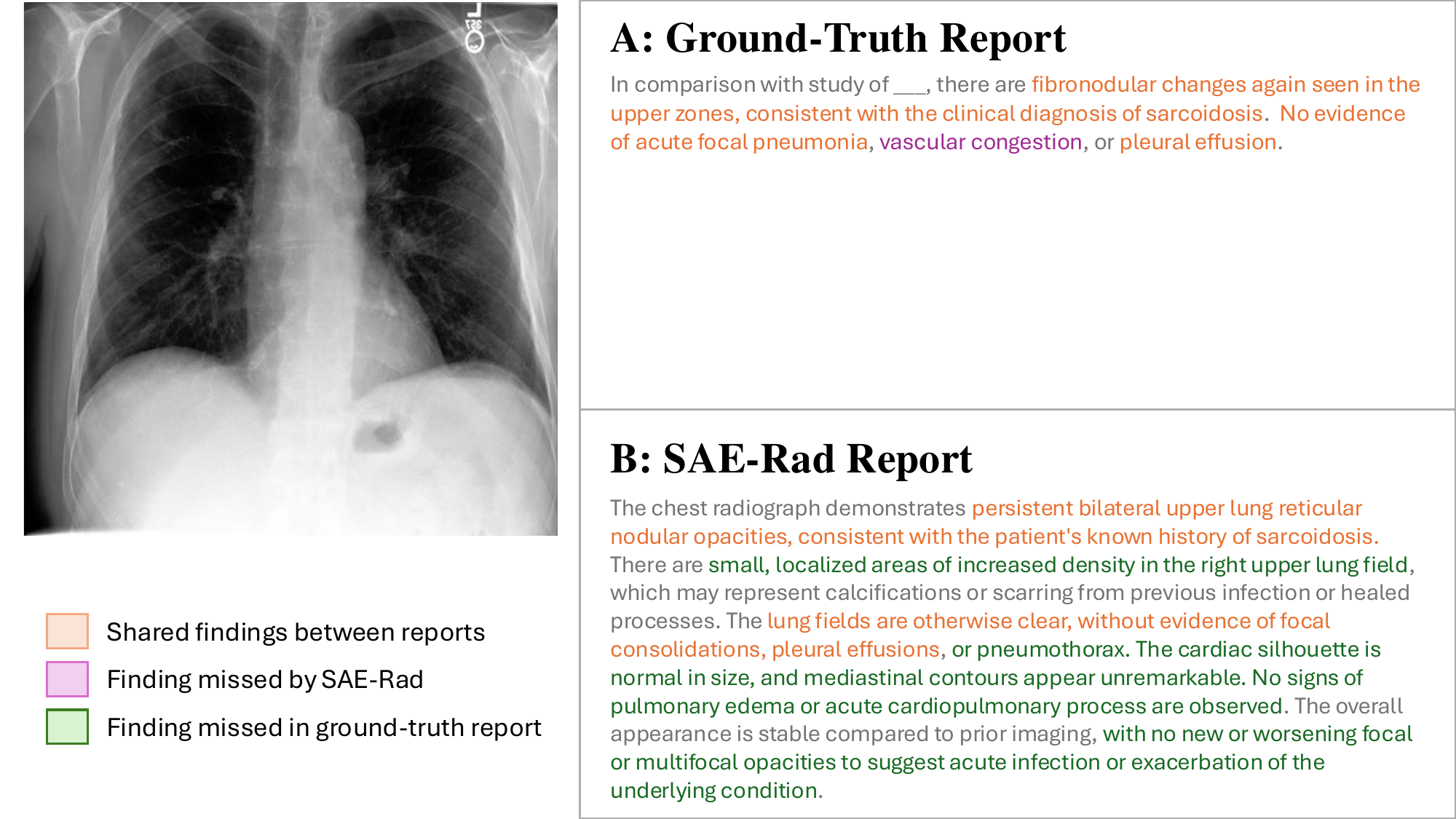}
    \caption{\textbf{Example image for which an SAE-Rad generated report has significantly more detail than the ground-truth.} SAE-Rad captures nearly all features noted in the ground-truth such as: fibronodular changes, a diagnosis of sarcoidosis and pleural effusions. However SAE-Rad also notes the lack of pneuomothorax, pulmonary edema and the increased density in the right upper lung field. Despite the clinical relevance of these elements this report likely scored poorly on RadFact due to the incompleteness of the ground-truth.}
    \label{radfact_limits} 
\end{figure}

\section{Ablation experimental setup}
\label{sec:appendix_h}

In this section we present the experimental setup for our ablation studies. We evaluated six SAEs on RadFact across three expansion factors and two sparsity levels. We evaluated the SAEs using a set of 60 randomly selected radiographs from the MIMIC-CXR test split. RadFact was evaluated using the ``Llama3-70B-Instruct'' model.
In this study, we used an earlier prompting template to generate feature explanations and subsequent reports.
Overall, we note that this change resulted in slightly worse quantitative metrics than those that were obtained using the final prompting strategy.

All SAEs were trained for 200,000 steps with a batch size of 2048. Models were trained using an Adam optimizer with no weight decay and a learning rate of \num{5e-5}, as well as a linear warm-up of the learning rate for the first 1\% of training, linear warm-down of learning rate for the last 20\% of training, and an L1 coefficient warm-up for the first 5\% of training.

We summarise experimental differences between runs in \cref{tab:sae-parameters}, as well as resultant L0 and explained variance metrics. The L0 and explained variance metrics in \cref{tab:sae-parameters} provide insights into the quality of our SAEs. Notably, the ``dense'' variants exhibit higher L0 values, indicating reduced sparsity, but achieve greater explained variance. The different variants therefore represent different points on the Pareto frontier between sparsity and reconstruction quality.

\begin{table}[htbp]
    \centering
    \caption{Varying hyperparameters and resulting metrics for the six SAEs.}
    \label{tab:sae-parameters}
    \begin{tabular}{l S[table-format=3] S[table-format=1.2e1] c c}
        \toprule
        & {Expansion factor} & {L1 coefficient} & L0 $\downarrow$ & Explained variance (\%) $\uparrow$ \\ 
        \midrule
        \texttimes128        & 128   & 6e-3      & 12.0  & 90.9 \\ 
        \texttimes64         & 64    & 8e-3      & 13.6  & 84.3 \\ 
        \texttimes32         & 32    & 9e-3      & 15.1  & 79.2 \\ 
        \texttimes128\_dense & 128   & 3.75e-3   & 26.0  & 93.2 \\ 
        \texttimes64\_dense  & 64    & 5e-3      & 28.4  & 88.5 \\ 
        \texttimes32\_dense  & 32    & 5.63e-3   & 31.8  & 84.7 \\
        \bottomrule
    \end{tabular}
\end{table}

\end{document}